\documentclass{article} % For LaTeX2e
\usepackage{iclr2027_conference,times}

\usepackage{amsmath,amsfonts,bm}

\def\eqref#1{equation~\ref{#1}}
\def\1{\bm{1}}

\DeclareMathAlphabet{\mathsfit}{\encodingdefault}{\sfdefault}{m}{sl}
\SetMathAlphabet{\mathsfit}{bold}{\encodingdefault}{\sfdefault}{bx}{n}

\usepackage{hyperref}
\usepackage[capitalize,noabbrev]{cleveref}
\usepackage{url}
\usepackage{booktabs}
\usepackage{graphicx}
\usepackage[table]{xcolor}
\usepackage{array}
\usepackage{caption}
\usepackage{amsmath}
\usepackage{dsfont}
\usepackage{algorithm}
\usepackage{algpseudocode}
\usepackage{wrapfig}
\newtheorem{proposition}{Proposition}
\newcommand{\ymean}[2]{#1{\scriptsize$\pm$#2}}
\newcommand{\ybest}[2]{\textbf{#1}{\scriptsize$\pm$#2}}
\makeatletter
\title{You Only Reprogram Once: Rethinking Prolonged Training for Visual Reprogramming}

\iclrfinalcopy

\author{
Zizhao Li$^{1}$,
Mohammed Yaqoob Ansari$^{1}$,
Xinyu Su$^{2}$,
Jiayang Ao$^{1}$,
Joseph West$^{1}$, \\
~\textbf{Kourosh Khoshelham}$^{1}$
\\
$^{1}$The University of Melbourne
$^{2}$Fudan University
}

\begin{document}

\maketitle

\begin{abstract}
Visual reprogramming is a parameter-efficient method for adapting pretrained models, yet its training can remain computationally expensive: even with a frozen backbone, visual prompts are often optimized through the full model for hundreds of epochs. Before changing what the pretrained model sees, we ask whether we are fully using what it already tells us. We find that modeling the full source response can already yield strong downstream predictions without prompt optimization. Motivated by this observation, we introduce You Only Reprogram Once (YORO), which constructs a downstream predictor from the frozen response space in a single forward-only traversal. Its Bayesian Discriminant Mapping (BDM) derives a covariance-aware affine mapping from streaming class statistics, requiring no backpropagation, optimizer updates, or repeated visits to the training set. When further input adaptation helps, YORO-FP optionally refines the visual prompt for 20 epochs. BDM also extends naturally to CLIP by treating attribute-prompt similarities as source responses. Across three full-data settings, YORO improves average accuracy over the strongest prior gradient-free mapping by 18.4--24.4\%. On 16-shot CLIP, it raises the four-backbone average from 71.4\% to 77.2\%. YORO-FP provides further gains on selected tasks, while validation often retains the one-pass predictor. These results suggest a different default for visual reprogramming: read out the frozen response first, and optimize the input only when needed. %\footnote{Code will be publicly available after the paper is accepted.}
\end{abstract}

\section{Introduction}
\label{sec:introduction}

Visual reprogramming adapts a pretrained model to a downstream task by
modifying lightweight input and output interfaces while keeping the backbone
frozen
\citep{elsayed2018adversarial,tsai2020transfer,chen2023understanding,
tsao2024autovp}.
A visual prompt transforms downstream inputs into the source input space, while
a label mapping converts source responses into target predictions.
Despite optimizing few parameters, gradient-based visual reprogramming still
requires repeated backpropagation through the frozen backbone, and recent
methods commonly train for hundreds of epochs
\citep{cai2024bayesian,cai2024sample}.
This raises an important question:
\emph{does effective visual reprogramming actually require prolonged prompt
optimization?}

A key difficulty is that prompt optimization and label mapping are usually
evaluated together.
Poor initial performance may therefore reflect either insufficient downstream
information or an insufficient readout of information already present in the
source response.
RLM, FLM, and ILM construct one-to-one source--target correspondences
\citep{elsayed2018adversarial,tsai2020transfer,chen2023understanding},
while BLM and BLM+ use predicted source labels or selected probabilities
\citep{cai2024bayesian}.
However, these mappings still use only part of the source response and may
discard information that separates downstream classes.

\begin{figure}[t]
  \centering
  \includegraphics[width=\textwidth]{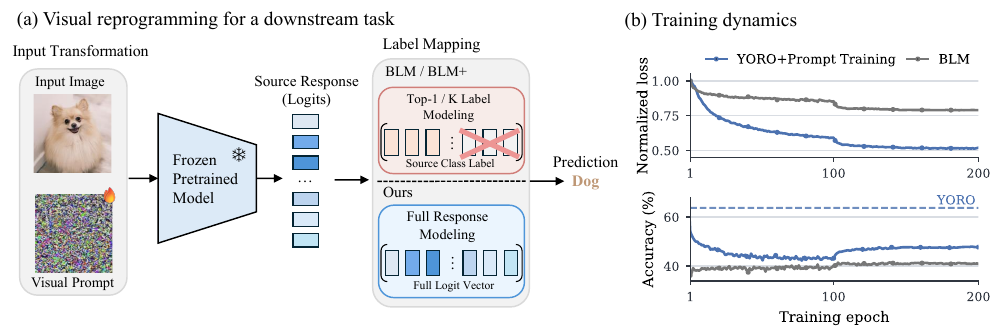}
  \vspace{-0.7cm}
  \caption{\textbf{Stronger label mapping, less prompt training.} Compared with \cite{cai2024bayesian}, which relies on Top-1 or Top-$K$
source-label information, our method (YORO) models the full source response. Using a single traversal of the
training set, YORO already outperforms BLM trained for 200 epochs on CIFAR-100, while further
visual-prompt optimization may reduce training loss without improving
validation accuracy.}
  \label{fig:training-does-not-help}
  \vspace{-0.2cm}
\end{figure}

Our central observation is that the output readout can change the apparent
need for prompt training. As illustrated in
\cref{fig:training-does-not-help}, prior label mapping methods model Top-1
or Top-$K$ source-label information, whereas our method models the full
source response. Figure~\ref{fig:training-does-not-help}(b) shows the case where, using the
same ResNet--18 backbone and the same watermark-style visual prompt
parameterization, a stronger readout of the initial frozen response already
outperforms a weaker mapping paired with prolonged prompt optimization. Moreover, continued prompt
training reduces the training loss without improving validation accuracy.
%This suggests that downstream structure should first be recovered from the frozen response space before modifying the input.

Motivated by this observation, we introduce
\emph{You Only Reprogram Once (YORO)}, a one-pass visual reprogramming
framework.
YORO fixes the input transformation at initialization, traverses the labeled
training set once using forward passes only, and constructs the downstream
predictor directly from the resulting responses.
It requires no backward pass, prompt update, intermediate features, or access
to model parameters, and can therefore operate on output-access black-box
models when full response vectors are available.

To make this one-pass formulation effective, we formulate
\emph{Bayesian Discriminant Mapping (BDM)}, an output label mapping method that
treats the complete response as a class-conditional observation. Following
Fisher's criterion \citep{fisher1936use}, target-class means encode separation
and a shared covariance measures within-class variation. The corresponding
Gaussian decision rule is affine, so its weights and biases directly form the
label mapping. To stabilize high-dimensional covariance estimation, BDM shrinks the
empirical covariance toward a trace-matched identity
\citep{friedman1989regularized}. BDM dynamically selects how much correlation
structure to retain, emphasizing stable discriminative directions without
amplifying noisy near-null directions. Streaming statistics recover the entire
regularized mapping in one traversal, without backpropagation or storing
individual responses.

This perspective also clarifies when additional prompt optimization can still
be useful.
If the initial frozen response does not contain sufficient downstream
structure, modifying the input can reshape the response space and improve the
readout.
We therefore introduce \emph{YORO-FP}, an optional few-pass extension that
starts from the one-pass YORO predictor and performs limited visual-prompt
refinement.
Rather than treating prolonged training as the default, YORO-FP makes input
adaptation an additional step whose value can be assessed against the
one-pass predictor.

Across three full-data backbone--input settings, YORO improves average
accuracy over the strongest prior gradient-free label mapping by
18.4--24.4\% and over the reported learned linear mapping by
6.3--12.3\%.
The same formulation extends naturally to CLIP by treating attribute-prompt
similarities as the source response.
In the 16-shot setting, YORO raises the four-backbone average accuracy from
71.4\% to 77.2\% while requiring no prompt optimization.
YORO-FP provides additional gains on selected tasks, although validation
frequently retains the original one-pass predictor.
These results establish YORO as a strong one-pass baseline for
visual reprogramming, showing that substantial downstream structure can be
recovered from the frozen response space before prompt optimization. Our contributions are:
\begin{itemize}
    \item We show that prolonged prompt optimization can be unnecessary when the frozen
response space is modeled effectively.

    \item We introduce BDM, a covariance-aware mapping that models the full
    source response and recovers an affine predictor in one pass.

    \item We introduce YORO, requiring one forward-only traversal and no prompt
    optimization, and YORO-FP for optional few-pass refinement.

    \item Across full-data visual reprogramming and few-shot CLIP adaptation,
    YORO substantially improves prior mappings and establishes a strong
    one-pass baseline.
\end{itemize}

\section{Related Work}
\label{sec:related_work}

\paragraph{Visual reprogramming and label mapping.}
Visual reprogramming adapts a frozen pretrained model through lightweight
input and output interfaces
\citep{elsayed2018adversarial,tsai2020transfer,chen2024model}.
Pixel-space methods typically learn either padding patterns
\citep{tsai2020transfer,chen2023understanding,tsao2024autovp}
or watermark-style perturbations
\citep{bahng2022exploring,cai2024sample,oh2023blackvip}.
Because source and downstream tasks generally use different label spaces,
an output label mapping is required to convert pretrained responses into
downstream predictions. RLM uses random source-to-target assignments
\citep{elsayed2018adversarial}, FLM selects frequently predicted source labels
\citep{tsai2020transfer}, and ILM iteratively updates this one-to-one mapping
during prompt training \citep{chen2023understanding}. BLM extends this design
to a dense probabilistic mapping from top-1 source predictions, while BLM+
incorporates top-$K$ predicted probabilities \citep{cai2024bayesian}.
Although these mappings can be constructed without gradient-based fitting,
the overall reprogramming pipelines still rely on multi-epoch optimization of
the input prompt.

Recent CLIP-based visual reprogramming instead improves image--text alignment.
AttrVR~\citep{cai2025attribute} introduces multiple attribute descriptions for
each downstream class, DVP~\citep{cai2025understanding} learns multiple visual
prompts with complementary roles, and Dual-Granularity Alignment
\citep{Wu_2026_CVPR} exploits semantic label hierarchies together with
multi-scale visual representations.

In contrast, BDM operates on the complete pretrained response rather than
compressed predictions or selected similarities, and models its downstream
class structure directly. Building on BDM, YORO turns this response geometry
into a streaming, one-pass reprogramming interface with
dataset-size-independent memory.

\paragraph{Training-efficient adaptation.}
Prior work mainly reduces or avoids prompt optimization. DAM-VP uses
meta-initialized cluster-specific prompts with 10-epoch adaptation
\citep{huang2023diversity}, AutoVP selects task-specific prompt, backbone, and
mapping configurations \citep{tsao2024autovp}, and LoR-VP adopts a low-rank
prompt parameterization \citep{jin2025lorvp}. Related
token-prompt methods improve initialization or optimization through gradient
regulation \citep{li2023gradient}, downstream prototypes
\citep{wang2024revisiting}, one-pass initialization \citep{park2025vipamin},
or stabilized prompt dynamics \citep{wang2026visual}. Other methods reduce
training further: CARPRT reweights prompt scores using
unlabeled target images \citep{dong2026carprt}. Tip-Adapter builds a few-shot feature cache for lightweight CLIP adaptation
\citep{zhang2022tipadapter}.  \citet{wang2024hard} apply Gaussian discriminant analysis directly in the
CLIP visual-feature space and therefore assume access to intermediate
features. In contrast, BDM operates
directly on model responses, allowing the same formulation to be applied to
both conventional classifiers and vision--language models
without requiring feature-space access. AReS instead reduces repeated model access by
using one service-API pass to prime a local encoder before reprogramming the
proxy \citep{zhang2026prime}. YORO goes further by constructing the downstream
predictor from labeled data in a single traversal, while YORO-FP tests whether
a small 20-epoch prompt budget provides additional value.

\section{Method}
\label{sec:method}

\subsection{Setting and the one-pass question}

\begin{figure}[t]
\centering
\includegraphics[width=\textwidth]{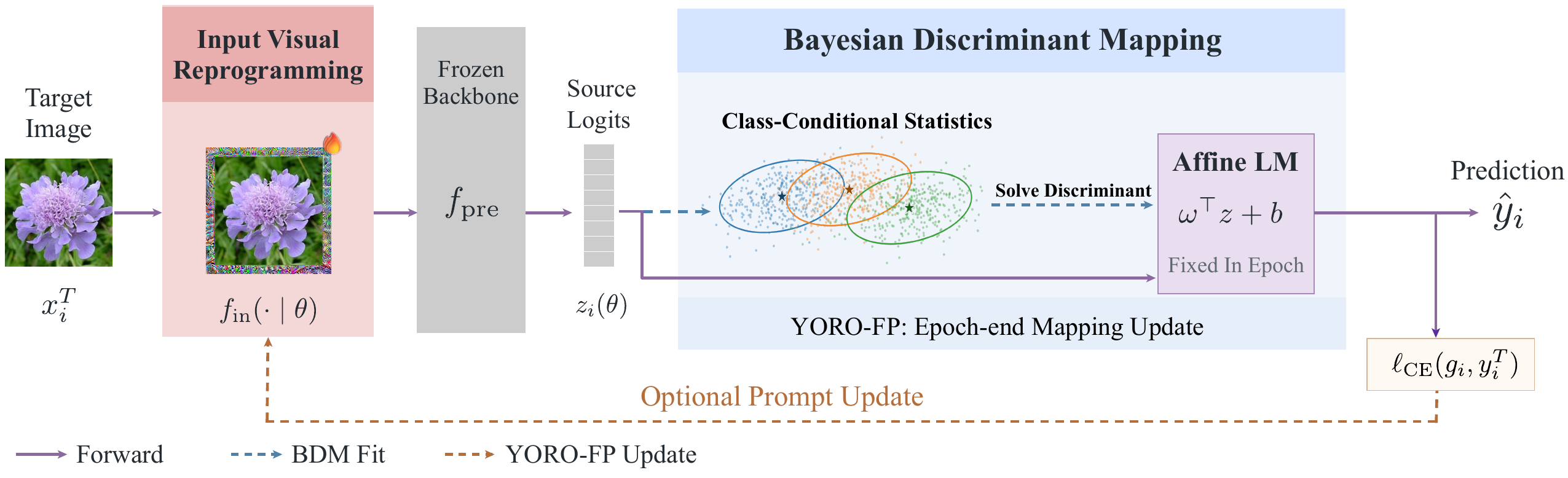}
\caption{\textbf{One pass first, optional prompt refinement when needed.}
YORO fits BDM to the full frozen response in one pass. BDM estimates target
class means and a shared covariance in the source-response space, then solves
these statistics into an affine label mapping. The response can be classifier
logits or CLIP image--text similarities. The dashed orange path denotes
optional YORO-FP refinement, where only the visual prompt is optimized and the
mapping is updated between epochs.}
\label{fig:yoro-framework}
\end{figure}

Following \citet{cai2024bayesian}, let the source and target label spaces be
$\mathcal Y^{\rm S}=\{1,\ldots,k_{\rm S}\}$ and
$\mathcal Y^{\rm T}=\{1,\ldots,k_{\rm T}\}$. A frozen source classifier
$f_{\rm pre}\colon\mathcal X^{\rm S}\rightarrow\mathbb R^{k_{\rm S}}$ and an
input transformation
$f_{\rm in}(\cdot\mid\theta)\colon\mathcal X^{\rm T}\rightarrow\mathcal X^{\rm S}$
produce
\begin{equation}
z_i(\theta)=(f_{\rm pre}\circ f_{\rm in})(x_i^{\rm T};\theta)
\in\mathbb R^{k_{\rm S}} .
\label{eq:source-logits}
\end{equation}
Here $\theta$ denotes the trainable parameters of the input visual prompt,
whereas $f_{\rm pre}$ remains frozen.
This response-space formulation also covers CLIP. Let
$\mathcal A=\{a_{c,j}:c\in[k_{\rm T}],j\in[m]\}$ be the attribute-prompt bank
used by DVP, with $m$ descriptions per downstream class. Given frozen CLIP
image and text encoders $\phi_{\rm I}$ and $\phi_{\rm T}$, we define
\begin{equation}
z_{i,c,j}(\theta)=\tau_{\rm C}
\left\langle
\bar\phi_{\rm I}(f_{\rm in}(x_i^{\rm T}\mid\theta)),
\bar\phi_{\rm T}(a_{c,j})
\right\rangle ,
\label{eq:clip-attribute-logits}
\end{equation}
where bars denote $\ell_2$ normalization and $\tau_{\rm C}$ is CLIP's frozen
logit scale \citep{radford2021learning,cai2025understanding}. Flattening
$(c,j)$ gives $z_i\in\mathbb R^{mk_{\rm T}}$, so we set
$k_{\rm S}=mk_{\rm T}$ and apply BDM without modification. The attribute
prompts and both CLIP encoders remain fixed. YORO-FP may update only the visual
input prompt. Since these logits lie in the visual
embedding subspace, our implementation uses an exact low-rank identity to
avoid forming the $(mk_{\rm T})\times(mk_{\rm T})$ covariance
(\cref{app:clip-low-rank}). It accumulates statistics and solves the system in
the visual embedding dimension while producing exactly the same logit-space
mapping.
We use an affine label mapping
\begin{equation}
f_{\rm out}^{\omega,b}(z)=\omega^\top z+b,\qquad
\omega\in\mathbb R^{k_{\rm S}\times k_{\rm T}},\quad
b\in\mathbb R^{k_{\rm T}} .
\label{eq:affine-lm}
\end{equation}
The conventional objective jointly adapts this interface through many visits
to the labeled target set. We instead first hold
$\theta=\theta_0$ fixed and ask whether $(\omega,b)$ can be constructed after
one visit to each training example.
The complete one-pass construction and its optional refinement path are
summarized in \cref{fig:yoro-framework}.

\subsection{Bayesian Discriminant Mapping}

Mathematically, BDM repurposes regularized linear discriminant analysis (LDA) \citep{fisher1936use,friedman1989regularized} as an output label mapping over the full pretrained response space. Existing mappings reduce the source response to labels,
selected probabilities, or predefined associations before predicting the
target class. BDM instead retains the complete response vector and asks which
response directions separate target classes relative to their within-class
variation. Target-class means describe the between-class geometry, while a
shared covariance provides the within-class metric and captures correlations
among response dimensions. A shared-covariance Gaussian model turns this
criterion into a Bayesian decision rule with the affine form in
\cref{eq:affine-lm}. Its parameters depend only on first- and second-order
statistics. We therefore model
\begin{equation}
Z\mid Y^{\rm T}=c\sim\mathcal N(\mu_c,\Sigma),\qquad
\pi_c=p(Y^{\rm T}=c).
\label{eq:gaussian-model}
\end{equation}
Here $\mu_c$ is the mean response for target class $c$, and $\Sigma$ is the
shared within-class response covariance.
The shared covariance supplies a common metric between class means and avoids
estimating a separate covariance for every target class. Up to terms
independent of $c$, Bayes' rule gives
\begin{equation}
g_c(z)=z^\top\Sigma^{-1}\mu_c
-\tfrac12\mu_c^\top\Sigma^{-1}\mu_c+\log\pi_c .
\label{eq:lda-score}
\end{equation}
Thus, the Gaussian decision rule is exactly an affine output mapping. Between
classes $c$ and $d$, its decision direction is
$\Sigma^{-1}(\mu_c-\mu_d)$, which emphasizes mean differences that are large
relative to within-class variation. Each covariance-adjusted mean
$\Sigma^{-1}\mu_c$ becomes the mapping vector for target class $c$, while the
bias encodes its class-conditional location and prior. This Fisher view is the
core of BDM: it converts the complete pretrained response geometry directly
into target logits, recovers the mapping from streaming statistics, and
requires no gradient-based optimization.

The key challenge is reliable covariance estimation. When $k_{\rm S}$ is large
relative to target supervision, especially in few-shot settings, the empirical
covariance can be noisy or singular. BDM therefore combines streaming
estimation with shrinkage. For each target class, it stores a count and
response sum together with one global second moment:
\begin{equation}
n_c=\sum_i\mathbf 1[y_i^{\rm T}=c],\qquad
s_c=\sum_{i:y_i^{\rm T}=c}z_i,\qquad
Q=\sum_i z_i z_i^\top .
\label{eq:streaming-statistics}
\end{equation}
Let $\mathbf n=(n_1,\ldots,n_{k_{\rm T}})$ and
$S=[s_1,\ldots,s_{k_{\rm T}}]$. After one traversal,
\begin{equation}
\widehat\mu_c=\frac{s_c}{n_c},\quad
\widehat\pi_c=\frac{n_c}{n},\quad
\widehat\Sigma=
\frac{Q-\sum_c s_cs_c^\top/n_c}{n-k_{\rm T}} .
\label{eq:streaming-estimates}
\end{equation}
To stabilize covariance estimation in the high-dimensional response space, BDM shrinks the covariance toward a trace-scaled identity:
\begin{equation}
\widehat\Sigma_\rho=(1-\rho)\widehat\Sigma+
\rho\frac{\operatorname{tr}(\widehat\Sigma)}{k_{\rm S}}I+\delta I ,
\qquad \rho\in[0,1],
\label{eq:shrinkage}
\end{equation}
where $\rho$ trades off empirical covariance structure against isotropic
stability, and $\delta\geq0$ is added only when required for stable
factorization. Shrinkage suppresses poorly estimated directions while
retaining informative covariance structure when supported by the data.

We select $\rho$ by validation accuracy. All candidates reuse the same
streaming statistics and cached validation responses, so the search requires
only one training traversal and one validation traversal. Candidate-specific
work is limited to forming $\widehat\Sigma_\rho$, solving the analytic output
mapping, and scoring the cached responses. Ties are assigned to the smaller
$\rho$.

Let $M=[\widehat\mu_1,\ldots,\widehat\mu_{k_{\rm T}}]$. BDM computes all target
class weights jointly:
\begin{equation}
\widehat\Sigma_\rho\omega_0=M,\qquad
b_{0,c}=-\tfrac12\widehat\mu_c^\top(\omega_0)_{:,c}
+\log\widehat\pi_c .
\label{eq:bdm-solve}
\end{equation}
The resulting predictor is
$f_{\rm out}^{\omega_0,b_0}(z)=\omega_0^\top z+b_0$.

\begin{proposition}[Exact streaming recovery]
\label{prop:streaming-equivalence}
For fixed responses $\{z_i\}$ and shrinkage $\rho$, the affine mapping
computed by BDM from the streaming statistics $(n,S,Q)$ is exactly the same
as that obtained by fitting BDM jointly to all training responses.
\end{proposition}

Thus, the one-pass formulation introduces no approximation: the class means
and shared covariance required by BDM are fully determined by $(n,S,Q)$.
YORO can therefore recover the exact BDM mapping in a single traversal,
without storing individual responses or revisiting the training set.
The proof is given in \cref{app:proof}.

\subsection{YORO-FP: controlled few-pass refinement}

BDM can recover target structure only when the frozen responses preserve it.
When they do not, the input interface can reshape the response space first.
YORO-FP therefore starts from
$(\theta_0,\omega_0,b_0)$ and permits 20 additional prompt epochs while
$f_{\rm pre}$ remains frozen. During epoch $t$, the active label mapping is fixed and
optimizer updates only $\theta$:
\begin{equation}
\mathcal L_{\rm FP}(\theta;\omega_t,b_t)=
\frac1n\sum_i\ell_{\rm CE}\!\left(
\omega_t^\top z_i(\theta)+b_t,y_i^{\rm T}\right).
\label{eq:yoro-fp-loss}
\end{equation}
The same minibatch logits, detached before the prompt update, accumulate a new
set of BDM statistics. At the epoch boundary these statistics yield a
candidate $(\widehat\omega_{t+1},\widehat b_{t+1})$. By default, we use the Momentum update to stabilize the LM as prompt
optimization gradually changes the response geometry. At each epoch, the
newly estimated BDM parameters are blended with the previous mapping:
\begin{equation}
\omega_{t+1}=\beta\omega_t+(1-\beta)\widehat\omega_{t+1},\qquad
b_{t+1}=\beta b_t+(1-\beta)\widehat b_{t+1},
\label{eq:momentum}
\end{equation}
with the fixed smoothing coefficient $\beta=0.9$. This smooths epoch-to-epoch changes in the discriminant
mapping while allowing it to track the evolving response space, without any
additional refitting traversal. \Cref{alg:yoro} in the appendix summarizes the shared
one-pass initialization and optional few-pass refinement.

\section{Experiments}
\label{sec:experiments}

\paragraph{Protocols.}
We evaluate YORO in two complementary settings. For full-data visual
reprogramming, we use
ImageNet-1K-pretrained ResNet--18
\citep{resnet,russakovsky2015imagenet} and Instagram-pretrained
ResNeXt--101--32x8d \citep{resnext}. Both expose 1,000 source logits. We evaluate on
Flowers102 \citep{flowers102}, DTD \citep{DTD}, UCF101 \citep{UCF101},
Food101 \citep{Food101}, GTSRB \citep{GTSRB}, EuroSAT \citep{EuroSAT},
OxfordPets \citep{Oxfordpets}, StanfordCars \citep{stanfordcars}, SUN397
\citep{sun397}, CIFAR-10/100 \citep{cifar10}, and SVHN \citep{SVHN}.
The watermark setting uses the same 12 tasks with ResNet--18. We compare
against RLM, FLM, ILM, BLM, BLM+
\citep{elsayed2018adversarial,tsai2020transfer,chen2023understanding,
cai2024bayesian}, and Deep. Deep is a gradient-based label mapping that uses
the same frozen backbone and input interface, but learns a dense single-layer
linear output mapping jointly with the visual prompt. All baselines follow
their original 200-epoch training protocols.

For few-shot adaptation, we follow the 16-shot DVP protocol
\citep{cai2025understanding}, covering 11 datasets and four frozen CLIP
backbones \citep{radford2021learning}. By default, DVP refers to the variant using unsupervised prompt clustering. YORO uses the same DVP attribute-prompt bank and treats the complete
image--text similarity vector in \cref{eq:clip-attribute-logits} as its frozen
response.

Unless stated otherwise, we report top-1 classification accuracy (\%) and
mean $\pm$ sample standard deviation over three random seeds. We consider two
input reprogramming forms: \emph{padding} places a learnable prompt around the
resized target image, whereas \emph{watermarking} overlays a learnable
perturbation on the image itself. Following \cite{cai2025attribute,cai2025understanding}, we use 16-pixel padding. YORO-FP performs 20 additional prompt epochs.
Further implementation details are provided in \cref{app:implementation}.

\begin{table}[t]
\captionsetup{position=top}
\caption{Performance comparison of visual reprogramming methods with different
output label mappings under the padding visual prompt setting (top-1 accuracy,
\%). {YORO and YORO-FP report mean $\pm$ sample standard deviation over three
seeds.} Baseline results are directly taken from
\citet{cai2024bayesian}. YORO results are highlighted, the best prior gradient-free label mapping is
shown in \textbf{bold}, and the gradient-based method (Deep) is shown in gray.
Our methods achieve the best overall results.}
\vspace{-0.2cm}
\label{tab:padding-detailed}
\centering
\small
\resizebox{\textwidth}{!}{%
\begin{tabular}{
c|cccccc|
>{\columncolor{blue!10}}c
>{\columncolor{blue!20}}c|
ccccc|
>{\columncolor{blue!10}}c
>{\columncolor{blue!20}}c
}
\toprule
& \multicolumn{8}{c|}{ResNet--18 (ImageNet--1K)}
& \multicolumn{7}{c}{ResNeXt--101--32x8d (Instagram)} \\
\midrule
Dataset
& RLM & FLM & ILM & BLM & BLM+
& {\color{gray!70}Deep}
& YORO & YORO-FP
& FLM & ILM & BLM & BLM+
& {\color{gray!70}Deep}
& YORO & YORO-FP \\
\midrule
Flowers102
& 11.0 & 20.0 & 27.9 & 44.4 & 50.1
& {\color{gray!70}76.7}
& \ybest{88.0}{0.0} & \ymean{87.9}{0.3}
& 22.5 & 27.9 & 31.5 & 30.1
& {\color{gray!70}85.2}
& \ymean{94.0}{0.0} & \ybest{94.3}{0.1} \\
DTD
& 16.3 & 32.4 & 35.3 & 42.0 & 43.9
& {\color{gray!70}49.1}
& \ymean{59.0}{0.0} & \ybest{59.6}{0.5}
& 40.3 & 41.4 & 47.8 & 49.4
& {\color{gray!70}64.0}
& \ybest{70.3}{0.0} & \ybest{70.3}{0.0} \\
UCF101
& 6.6 & 18.9 & 23.9 & 30.9 & 32.0
& {\color{gray!70}46.0}
& \ybest{61.3}{0.0} & \ybest{61.3}{0.0}
& 41.9 & 43.1 & 48.3 & 50.1
& {\color{gray!70}68.3}
& \ymean{74.9}{0.0} & \ybest{75.4}{0.3} \\
Food101
& 3.8 & 12.8 & 14.8 & 23.2 & 25.1
& {\color{gray!70}34.1}
& \ybest{49.8}{0.0} & \ybest{49.8}{0.0}
& 20.5 & 23.0 & 29.6 & 31.4
& {\color{gray!70}58.7}
& \ybest{70.3}{0.0} & \ybest{70.3}{0.0} \\
GTSRB
& 46.1 & 45.5 & 52.0 & 54.8 & 54.3
& {\color{gray!70}63.1}
& \ymean{79.0}{0.1} & \ybest{80.2}{0.4}
& 56.2 & 59.9 & 62.9 & 63.0
& {\color{gray!70}74.4}
& \ymean{79.3}{0.1} & \ybest{79.6}{0.5} \\
EuroSAT
& 82.4 & 83.8 & 85.2 & 86.7 & 86.7
& {\color{gray!70}92.4}
& \ybest{92.8}{0.0} & \ymean{92.5}{0.2}
& 87.8 & 86.2 & 87.6 & 88.3
& {\color{gray!70}93.2}
& \ymean{93.0}{0.0} & \ybest{93.4}{0.3} \\
OxfordPets
& 9.3 & 62.9 & 65.4 & 69.8 & 70.6
& {\color{gray!70}73.0}
& \ybest{86.9}{0.0} & \ybest{86.9}{0.0}
& 76.8 & 78.9 & 82.4 & 83.0
& {\color{gray!70}91.8}
& \ymean{94.8}{0.0} & \ybest{95.1}{0.1} \\
StanfordCars
& 0.9 & 2.7 & 4.5 & 5.4 & 7.7
& {\color{gray!70}14.3}
& \ymean{34.9}{0.0} & \ybest{37.1}{0.1}
& 4.6 & 7.0 & 8.3 & 9.3
& {\color{gray!70}50.5}
& \ymean{71.1}{0.0} & \ybest{72.2}{0.3} \\
SUN397
& 1.0 & 10.4 & 13.0 & 16.2 & 18.7
& {\color{gray!70}26.3}
& \ybest{47.8}{0.0} & \ybest{47.8}{0.0}
& 21.6 & 23.7 & 30.1 & 32.0
& {\color{gray!70}51.5}
& \ybest{61.9}{0.0} & \ybest{61.9}{0.0} \\
CIFAR10
& 63.0 & 65.7 & 65.5 & 66.7 & 66.8
& {\color{gray!70}72.1}
& \ybest{86.2}{0.1} & \ybest{86.2}{0.1}
& 80.3 & 81.7 & 82.2 & 82.2
& {\color{gray!70}83.4}
& \ybest{95.3}{0.0} & \ybest{95.3}{0.0} \\
CIFAR100
& 12.9 & 18.1 & 24.8 & 29.6 & 30.6
& {\color{gray!70}46.7}
& \ybest{64.8}{0.1} & \ybest{64.8}{0.1}
& 39.7 & 45.9 & 47.8 & 47.8
& {\color{gray!70}56.2}
& \ybest{76.3}{0.1} & \ybest{76.3}{0.1} \\
SVHN
& 73.5 & 73.1 & 75.2 & 74.5 & 74.2
& {\color{gray!70}82.1}
& \ymean{72.3}{0.1} & \ybest{77.0}{0.3}
& 79.0 & \textbf{81.4} & 79.8 & 79.3
& {\color{gray!70}85.7}
& \ymean{56.5}{0.0} & \ymean{68.6}{0.6} \\
\midrule
Average
& 27.2 & 37.2 & 40.6 & 45.3 & 46.7
& {\color{gray!70}56.3}
& \ymean{68.6}{0.0} & \ybest{69.3}{0.1}
& 47.6 & 50.0 & 53.2 & 53.8
& {\color{gray!70}71.9}
& \ymean{78.2}{0.0} & \ybest{79.4}{0.1} \\
\bottomrule
\end{tabular}}
\vspace{-0.1cm}
\end{table}

\begin{wraptable}{r}{0.5\linewidth}
\vspace{-1.2cm}
\centering
\small
\setlength{\tabcolsep}{4pt}
\renewcommand{\arraystretch}{1.08}
\caption{Watermarking-based visual reprogramming on ResNet--18 (top-1
accuracy, \%). Baseline results are taken from \cite{cai2024bayesian}.
Our methods achieve the best overall results.}
\label{tab:watermark-yoro}
\resizebox{0.5\textwidth}{!}{%
\begin{tabular}{l|ccc|c|>{\columncolor{blue!10}}c>{\columncolor{blue!20}}c}
\toprule
& \multicolumn{6}{c}{ResNet--18 (ImageNet--1K)} \\
\cmidrule(lr){2-7}
Dataset & ILM & BLM & BLM+ & {\color{gray!70}Deep}
& YORO & YORO-FP \\
\midrule
Flowers102   & 23.2 & 39.2 & 44.1 & {\color{gray!70}82.4} & \ybest{88.8}{0.0} & \ybest{88.8}{0.0} \\
DTD          & 29.0 & 40.1 & 43.0 & {\color{gray!70}48.9} & \ybest{62.5}{0.0} & \ybest{62.5}{0.0} \\
UCF101       & 24.4 & 32.9 & 35.4 & {\color{gray!70}53.1} & \ybest{65.1}{0.0} & \ybest{65.1}{0.0} \\
Food101      & 13.2 & 21.5 & 22.9 & {\color{gray!70}30.4} & \ybest{52.8}{0.0} & \ybest{52.8}{0.0} \\
GTSRB        & 76.8 & 82.1 & 82.0 & {\color{gray!70}89.5} & \ymean{75.9}{0.1} & \ybest{84.5}{0.6} \\
EuroSAT      & 84.3 & 84.4 & 84.8 & {\color{gray!70}89.2} & \ybest{93.2}{0.0} & \ybest{93.2}{0.0} \\
OxfordPets   & 70.0 & 72.4 & 73.3 & {\color{gray!70}77.6} & \ybest{88.5}{0.0} & \ybest{88.5}{0.0} \\
StanfordCars &  3.4 &  5.5 &  7.4 & {\color{gray!70}30.7} & \ymean{37.8}{0.0} & \ybest{38.6}{0.5} \\
SUN397       & 13.4 & 18.4 & 19.4 & {\color{gray!70}32.9} & \ybest{52.0}{0.0} & \ybest{52.0}{0.0} \\
CIFAR10      & 68.9 & 74.9 & 75.7 & {\color{gray!70}71.7} & \ybest{86.9}{0.1} & \ybest{86.9}{0.1} \\
CIFAR100     & 33.8 & 41.2 & 41.6 & {\color{gray!70}39.9} & \ybest{64.8}{0.1} & \ybest{64.8}{0.1} \\
SVHN         & 78.3 & 79.2 & 78.8 & {\color{gray!70}83.7} & \ymean{60.6}{0.1} & \ybest{84.7}{0.2} \\
\midrule
Average      & 43.2 & 49.3 & 50.7 & {\color{gray!70}60.8} & \ymean{69.1}{0.0} & \ybest{71.9}{0.0} \\
\bottomrule
\end{tabular}}
\vspace{-0.5cm}
\end{wraptable}

\subsection{Results on Image Classifiers}

\Cref{tab:padding-detailed,tab:watermark-yoro} show that YORO outperforms
prior gradient-free label mappings on average across all three settings,
without iterative prompt optimization. The gains span diverse tasks,
including StanfordCars, CIFAR-100, UCF101, Food101, OxfordPets, and SUN397.

For padding, we follow \citet{cai2025attribute,cai2025understanding} and
resize images to $192\times192$ within a $224\times224$ canvas, leaving
a 16-pixel border. This standardizes the image area across datasets,
avoids excessively small image regions under legacy padding, and limits
the prompt to 39,936 trainable parameters during optional refinement.
We also evaluate YORO under the legacy configuration, which applies no
resizing and instead uses dynamic padding to match the pretrained input size,
as detailed in \Cref{sec:visual-prompt-config}. YORO still substantially outperforms BLM+
on both backbones, demonstrating gains from the output mapping without
the change in input resizing.

For watermarking, all methods resize images to the pretrained model's input
resolution before applying the visual prompt. Under this matched protocol,
YORO achieves 69.1\% average accuracy on ResNet--18, compared with 50.7\%
for BLM+ and 60.8\% for Deep, without iterative prompt optimization.
YORO-FP further reaches 71.9\%, with the largest gains on GTSRB and SVHN,
where input adaptation is more useful.

These gains also come at substantially lower adaptation cost. While the
baselines optimize visual prompts for 200 epochs, YORO requires only one
forward-only traversal and an analytic mapping solve. YORO-FP uses only
20 additional prompt-training epochs, around 10\% of the baseline training
budget.

%\WFclear

\subsection{Few-Shot Results on CLIP}

\begin{table}[t]
\captionsetup{position=top}
\caption{Performance comparison on 16-shot downstream classification using
ViT-B/32 and ResNet--50 CLIP backbones (accuracy \%). Prior results are taken from
\citet{cai2025understanding}. Highlighted YORO columns report mean
$\pm$ sample standard deviation over three seeds using DVP
attribute-prompt similarities as source responses. Best results are in
\textbf{bold}. Our methods achieve the best overall results.}
\vspace{-0.2cm}
\label{tab:fsclip-v32-r50}
\centering
\small
\resizebox{\textwidth}{!}{%
\begin{tabular}{
c|cccc
>{\columncolor{blue!10}}c
>{\columncolor{blue!20}}c|
cccc
>{\columncolor{blue!10}}c
>{\columncolor{blue!20}}c
}
\toprule
& \multicolumn{6}{c|}{ViT-B/32 (CLIP)}
& \multicolumn{6}{c}{ResNet--50 (CLIP)} \\
\midrule
Dataset
& VP & AR & AttrVR & DVP & YORO & YORO-FP
& VP & AR & AttrVR & DVP & YORO & YORO-FP \\
\midrule
FGVC Aircraft
& 24.3 & 21.8 & 24.5 & 26.1
& \ymean{36.6}{0.8} & \ybest{36.9}{1.2}
& 16.2 & 18.6 & 20.7 & 22.1
& \ymean{36.2}{1.5} & \ybest{36.3}{1.4} \\
Caltech101
& 92.3 & 92.7 & 92.0 & 92.9
& \ybest{94.0}{0.0} & \ymean{93.7}{0.6}
& 80.1 & 86.5 & 89.1 & 89.8
& \ybest{91.3}{0.1} & \ymean{91.0}{0.4} \\
StanfordCars
& 58.6 & 56.9 & 56.6 & 56.5
& \ybest{71.8}{0.6} & \ymean{71.0}{1.2}
& 44.0 & 53.9 & 53.9 & 54.5
& \ybest{68.6}{0.2} & \ymean{68.5}{0.3} \\
DTD
& 54.9 & 49.9 & 56.8 & 57.2
& \ybest{65.8}{1.6} & \ymean{65.5}{1.8}
& 43.4 & 46.4 & 54.4 & 55.9
& \ybest{64.4}{0.2} & \ymean{63.8}{0.8} \\
EuroSAT
& 85.9 & 85.6 & \textbf{88.6} & 88.5
& \ymean{82.3}{1.5} & \ymean{82.1}{2.4}
& 59.7 & 66.6 & 72.0 & 72.2
& \ybest{82.9}{0.7} & \ymean{79.7}{5.7} \\
Flowers102
& 71.2 & 66.7 & 77.8 & 82.5
& \ybest{94.0}{0.1} & \ybest{94.0}{0.1}
& 53.6 & 60.9 & 74.8 & 80.0
& \ybest{92.6}{0.5} & \ymean{92.5}{0.7} \\
Food101
& 75.0 & 75.7 & \textbf{77.2} & 77.0
& \ymean{75.1}{1.3} & \ymean{75.1}{1.3}
& 65.3 & 74.2 & \textbf{75.3} & 75.0
& \ymean{72.8}{0.2} & \ymean{72.6}{0.4} \\
OxfordPets
& 86.8 & 84.7 & \textbf{89.8} & 89.2
& \ymean{87.9}{1.0} & \ymean{87.5}{1.1}
& 77.2 & 82.5 & \textbf{88.9} & \textbf{88.9}
& \ymean{87.5}{0.9} & \ymean{87.5}{0.9} \\
SUN397
& 61.0 & 59.9 & 62.8 & 64.2
& \ybest{70.6}{0.2} & \ybest{70.6}{0.2}
& 48.8 & 56.8 & 59.9 & 61.1
& \ybest{67.6}{0.3} & \ybest{67.6}{0.3} \\
UCF101
& 67.3 & 63.5 & 67.9 & 70.5
& \ybest{77.8}{1.5} & \ybest{77.8}{1.5}
& 52.0 & 59.7 & 63.6 & 65.9
& \ybest{75.1}{0.2} & \ybest{75.1}{0.2} \\
RESISC45
& 73.9 & 71.6 & 73.9 & 76.0
& \ybest{82.6}{0.3} & \ymean{82.0}{0.5}
& 47.7 & 58.4 & 58.2 & 60.8
& \ybest{79.4}{1.1} & \ymean{78.6}{1.6} \\
\midrule
Average
& 68.3 & 66.3 & 69.8 & 71.0
& \ybest{76.2}{0.3} & \ymean{76.0}{0.2}
& 53.5 & 60.4 & 64.6 & 66.0
& \ybest{74.4}{0.2} & \ymean{73.9}{0.6} \\
\bottomrule
\end{tabular}}
%\vspace{-0.25cm}
\end{table}

\begin{wraptable}{r}{0.5\linewidth}
\vspace{-0.35cm}
\centering
\small
\caption{Aggregate 16-shot accuracy (\%) over 11 datasets. {YORO results
report mean $\pm$ sample standard deviation over three seeds.}
Overall averages the four frozen CLIP backbones equally. Our methods achieve the best overall results.}
\vspace{-0.2cm}
\label{tab:fs-summary}
\resizebox{0.5\textwidth}{!}{%
\begin{tabular}{lccccc}
\toprule
Method & RN50 & RN101 & ViT-B/32 & ViT-B/16 & Overall \\
\midrule
VP     & 53.5 & 57.5 & 68.3 & 74.4 & 63.4 \\
AR     & 60.4 & 62.7 & 66.3 & 76.5 & 66.5 \\
AttrVR & 64.6 & 67.2 & 69.8 & 78.5 & 70.0 \\
DVP    & 66.0 & 68.8 & 71.0 & 79.7 & 71.4 \\
\rowcolor{blue!10}
YORO   & \ybest{74.4}{0.2} & \ybest{76.8}{0.2}
       & \ybest{76.2}{0.3} & \ymean{81.2}{0.1} & \ybest{77.2}{0.1} \\
\rowcolor{blue!20}
YORO-FP  & \ymean{73.9}{0.6} & \ymean{76.7}{0.3}
       & \ymean{76.0}{0.2} & \ybest{81.4}{0.3} & \ymean{77.0}{0.3} \\
\bottomrule
\end{tabular}}
\vspace{-0.35cm}
\end{wraptable}

YORO can also be applied to CLIP-based few-shot learning. Following DVP
\citep{cai2025understanding}, we construct the source response from cosine
similarities between image embeddings and attribute prompts, and apply BDM to
learn the downstream class geometry in a single pass. Full per-dataset results
for ViT-B/32 and ResNet--50 are reported in
\cref{tab:fsclip-v32-r50}, with the four-backbone summary in
\cref{tab:fs-summary}. Across all four backbones, the highest average accuracy is achieved by
either YORO or YORO-FP.

The gains are particularly strong on the weaker backbones, reaching 8.4\% and
8.0\% on ResNet--50 and ResNet--101, compared with 5.2\% on ViT-B/32 and
1.5\% on ViT-B/16. At the dataset level, YORO provides large improvements on
fine-grained tasks such as FGVC Aircraft and StanfordCars, as well as
Flowers102, SUN397, UCF101, and RESISC45. For example, on ViT-B/32 it improves
DVP from 26.1\% to 36.6\% on FGVC Aircraft and from 56.5\% to 71.8\% on
StanfordCars. This pattern suggests that modeling the full similarity response
can recover class structure that is not captured by the original DVP readout,
especially when the frozen backbone provides less separable features.

YORO-FP provides little additional benefit on average. It improves the
backbone-level average only on ViT-B/16, while matching or slightly trailing
YORO on the other three backbones. This indicates that, for most CLIP settings,
the stronger full-response readout already captures most of the available
adaptation signal before prompt refinement, with additional input optimization
becoming useful only in selected cases.

\begin{figure}[t]
\centering
\includegraphics[width=\textwidth]{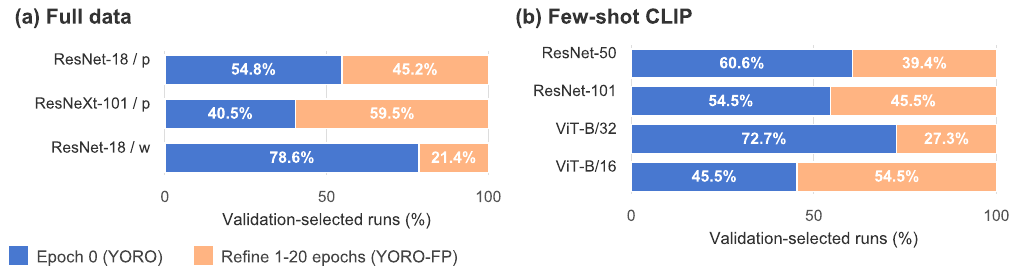}
\vspace{-0.2cm}
\caption{\textbf{Prompt refinement is beneficial only for a subset of runs.}
Fraction of three-seed runs whose best validation checkpoint is Epoch~0
(YORO) or Epochs~1--20 (YORO-FP) for (a) full-data visual reprogramming
and (b) 16-shot CLIP. Here, $p$ and $w$ denote padding and watermarking,
respectively.}
\label{fig:training-dynamics}
\end{figure}

\subsection{Does visual prompt training still help?}

\Cref{fig:training-dynamics} examines whether continued prompt refinement is
selected by validation. Across five of the seven backbone--setting
combinations, Epoch~0 is selected more often than refined checkpoints,
including 78.6\% of runs for ResNet--18 watermarking and 72.7\% for ViT-B/32
CLIP. Refinement is selected more often for ResNeXt--101 padding and ViT-B/16
CLIP, showing that additional input adaptation remains useful in some settings.

SVHN provides the clearest dataset-level example: YORO-FP improves over YORO
by 4.7\%, 12.1\%, and 24.1\% across the three full-data settings, with gains
also observed on GTSRB. These tasks differ substantially from the natural-image
categories used for source pretraining, making prompt refinement more useful
for reshaping the response space.

%Overall, prolonged prompt optimization should not be the default: many runs retain the one-pass YORO predictor, while YORO-FP is most useful when the initial frozen response space requires further adaptation.

\begin{wraptable}{r}{0.45\linewidth}
\vspace{-1cm}
\captionsetup{position=top}
\caption{{Accuracy and training time on Oxford Flowers with CLIP
ViT-B/32 under the 16-shot setting.} Results are mean $\pm$ sample standard
deviation over three seeds. YORO achieves the best accuracy with the lowest training cost. All methods in this table are rerun under the same hardware protocol.}
\label{tab:clip-efficiency}
\centering
\scriptsize
\setlength{\tabcolsep}{5pt}
\begin{tabular}{lcc}
\toprule
Method & Test accuracy (\%) & Training time (s) \\
\midrule
\rowcolor{blue!10}
YORO   & $\mathbf{94.0\pm0.1}$ & $\mathbf{17.1\pm6.2}$ \\
DGA    & $91.8\pm4.8$ & $6562.1\pm34.5$ \\
AutoVP & $81.5\pm0.4$ & $2754.4\pm21.3$ \\
DVP    & $80.9\pm1.1$ & $3714.9\pm7.3$ \\
AttrVR & $76.2\pm0.6$ & $2683.2\pm7.4$ \\
BLM    & $73.1\pm1.2$ & $4996.0\pm122.3$ \\
BLM+   & $72.7\pm1.0$ & $5019.9\pm20.9$ \\
\bottomrule
\end{tabular}
\vspace{-0.5cm}
\end{wraptable}

\subsection{Accuracy and Training Cost.}

The accuracy--efficiency advantage is clear on Oxford Flowers with CLIP
ViT-B/32 under the 16-shot setting (\cref{tab:clip-efficiency}). YORO achieves
the highest accuracy, 94.0\%, in only 17.1 seconds. DGA
\citep{Wu_2026_CVPR}, the closest baseline in accuracy, reaches 91.8\% but
requires 6562.1 seconds. Other methods show the same gap: DVP and AttrVR require
thousands of seconds while reaching 80.9\% and 76.2\%, respectively, while
AutoVP, BLM, and BLM+ achieve only 72.7--81.5\% with 2754.4--5019.9 seconds
of training.

YORO therefore improves efficiency without sacrificing accuracy. Instead of
repeatedly optimizing the visual prompt through the frozen backbone, it
constructs the downstream mapping from a single forward traversal of the
labeled data. This supports our central hypothesis that prolonged prompt
optimization is often unnecessary when the frozen response space can already
be read out effectively.

\subsection{Ablation Study}

We study how the affine label mapping should be handled during the optional
YORO-FP prompt refinement. As the visual prompt is updated, the frozen
backbone produces a gradually changing response space, raising the question of
whether the initial YORO mapping should remain fixed or adapt to this change.
We compare three strategies. \textbf{Frozen} keeps the initial YORO mapping
$(\omega_0,b_0)$ unchanged throughout refinement. \textbf{Refresh} directly replaces the mapping with the BDM estimate
accumulated during the current prompt epoch. \textbf{Momentum} instead smooths each
new mapping with the previous one using \cref{eq:momentum} with $\beta=0.9$.
Equivalently, Frozen and Refresh correspond to $\beta=1$ and $\beta=0$,
respectively. By default, we use the \textbf{Momentum} update strategy.

The two ablations use task suites tailored to their protocols. The full-data
study covers 14 datasets: the 12 datasets in the published-baseline comparison
plus Caltech101 \citep{fei2004caltech101} and RESISC45
\citep{cheng2017resisc45}. The few-shot study follows DVP and averages over
its 11 datasets and four CLIP backbones.

\Cref{tab:yoro-mapping-full} shows that Momentum performs best in all three
full-data settings. Compared with one-pass YORO, it improves average accuracy
by 0.6\% with ResNet--18 padding, 2.4\% with ResNet--18 watermarking, and
1.1\% with ResNeXt--101 padding. Refresh generally improves over Frozen,
indicating that adapting the mapping to the evolving response space can be
useful, while Momentum provides a further benefit by avoiding abrupt
epoch-to-epoch changes. The larger gain under watermarking is concentrated in
tasks such as SVHN and GTSRB, where prompt adaptation changes the response
space more substantially. Per-dataset results are reported in
\cref{tab:yoro-mapping-full-detail}.

The few-shot CLIP results in \cref{tab:yoro-mapping-fsclip} show a similar but
weaker effect. Momentum is the strongest YORO-FP update rule overall and gives
the best result on ViT-B/16, but one-pass YORO remains best when averaged
across all four backbones, with 77.2\% accuracy. Thus, smoothing the mapping is
useful when prompt refinement meaningfully changes the response space, while
the broader result remains unchanged: in most settings, the initial one-pass
mapping already captures most of the available downstream structure.

\providecommand{\abacc}[2]{#1{\scriptsize$\pm$#2}}
\providecommand{\abest}[2]{\textbf{#1}{\scriptsize$\pm$#2}}

\begin{table}[t]
\centering
\begin{minipage}[t]{0.485\textwidth}
\vspace{0pt}
\captionsetup{position=top}
\caption{\textbf{Full-data mapping-update ablation} (top-1 accuracy, \%).
Entries are 14-dataset macro-averages over three seeds.
YORO is the one-pass predictor, while Frozen, Refresh, and Momentum denote
YORO-FP variants with $\beta=1$, $0$, and $0.9$, respectively.
P/W denote padding/watermarking. }
\vspace{-0.15cm}
\label{tab:yoro-mapping-full}
\centering
\footnotesize
\setlength{\tabcolsep}{1.5pt}
\begin{tabular}{@{}lcccc@{}}
\toprule
Setting & YORO & Frozen & Refresh & Momentum \\
\midrule
RN18--P    & \abacc{71.0}{0.0} & \abacc{71.1}{0.0} & \abacc{71.1}{0.0} & \abest{71.6}{0.1} \\
RN18--W    & \abacc{71.5}{0.0} & \abacc{73.3}{0.1} & \abacc{73.6}{0.1} & \abest{73.9}{0.0} \\
RNXT101--P & \abacc{79.8}{0.0} & \abacc{80.2}{0.0} & \abacc{80.6}{0.1} & \abest{80.9}{0.1} \\
\midrule
Overall    & \abacc{74.1}{0.0} & \abacc{74.9}{0.0} & \abacc{75.1}{0.1} & \abest{75.4}{0.0} \\
\bottomrule
\end{tabular}
\vspace{-0.15cm}
\end{minipage}
\hfill
\begin{minipage}[t]{0.485\textwidth}
\vspace{0pt}
\captionsetup{position=top}
\caption{\textbf{Few-shot CLIP mapping-update ablation} (top-1 accuracy, \%).
Entries are 11-dataset macro-averages over three seeds.
YORO is the one-pass predictor, while Frozen, Refresh, and Momentum denote
YORO-FP variants with $\beta=1$, $0$, and $0.9$, respectively.}
\label{tab:yoro-mapping-fsclip}
\centering
\footnotesize
\setlength{\tabcolsep}{1.5pt}
\begin{tabular}{@{}lcccc@{}}
\toprule
Backbone & YORO & Frozen & Refresh & Momentum \\
\midrule
ResNet--50  & \abest{74.4}{0.2} & \abacc{74.0}{0.2} & \abacc{73.8}{0.4} & \abacc{73.9}{0.6} \\
ResNet--101 & \abest{76.8}{0.2} & \abacc{76.5}{0.2} & \abacc{76.4}{0.0} & \abacc{76.7}{0.3} \\
ViT-B/32    & \abest{76.2}{0.3} & \abacc{76.0}{0.4} & \abacc{76.1}{0.2} & \abacc{76.0}{0.2} \\
ViT-B/16    & \abacc{81.2}{0.1} & \abacc{80.9}{0.2} & \abacc{81.3}{0.3} & \abest{81.4}{0.3} \\
\midrule
Overall     & \abest{77.2}{0.1} & \abacc{76.8}{0.1} & \abacc{76.9}{0.1} & \abacc{77.0}{0.3} \\
\bottomrule
\end{tabular}
\end{minipage}
%\vspace{-0.15cm}
\end{table}

\section{Limitations}
\label{sec:limitations}

YORO assumes that the frozen response space retains useful structure for the
target labels. SVHN shows that this assumption may weaken when the target label
space differs substantially from the source label space, in which case visual
prompt adaptation remains useful for reshaping the responses. Our experiments
also focus on image classification. Extending the one-pass principle to other tasks and modalities
remains an interesting direction.

\section{Conclusion}
\label{sec:conclusion}

Visual reprogramming often adopts long optimization schedules without first
establishing whether they are necessary. YORO shows that a single labeled-data
traversal can already produce a strong downstream predictor when the frozen
response space is read out effectively. Using Bayesian Discriminant Mapping,
YORO requires no backward pass or prompt update while consistently improving
prior label mappings across full-data visual reprogramming and few-shot CLIP.

These results establish YORO as a strong one-pass baseline. They suggest that
future methods should first determine how much downstream structure is already
recoverable from frozen responses before introducing further prompt
optimization. YORO-FP covers the complementary case where additional input
adaptation is useful. This one-pass-to-few-pass perspective provides a simple
reference for more efficient visual reprogramming.

\bibliography{iclr2027_conference}
\bibliographystyle{iclr2027_conference}

\appendix
\newpage
\section{Implementation and evaluation details}
\label[appendix]{app:implementation}

\paragraph{Full-data protocol.}
Following AttrVR and DVP \citep{cai2025attribute,cai2025understanding}, we use
the same padding for the principal experiment: a downstream image is
resized to $192\times192$ and embedded in the $224\times224$ source canvas.
This parameterization contains 39,936 trainable RGB values, substantially
fewer than larger padding designs. The additive
watermark is initially zero. The initial BDM pass is deterministic. YORO-FP
uses random resized crops with scale $[0.7,1.0]$ and ratio $[0.9,1.1]$, plus
horizontal flips except for GTSRB and SVHN. We use batch size 256 for
ResNet--18 and 64 for ResNeXt. Adam optimizer~\citep{kingma2015adam} updates only the prompt with learning rate
0.01, default $(\beta_1,\beta_2)=(0.9,0.999)$, no weight decay, and no
scheduler.
The additional watermark experiment overlays a trainable input-space pattern.
Full-data training-dynamics and update ablations include Caltech101
\citep{fei2004caltech101} and RESISC45 \citep{cheng2017resisc45} in
addition to the 12 tasks in the published-baseline comparison. All experiments use three random seeds and the configurations specified above.

\paragraph{Few-shot CLIP protocol.}
The 16-shot experiments use FGVC Aircraft,
Caltech101, StanfordCars, DTD, EuroSAT, Flowers102, Food101, OxfordPets,
SUN397, UCF101, and RESISC45
\citep{maji2013aircraft,fei2004caltech101,stanfordcars,DTD,EuroSAT,
flowers102,Food101,Oxfordpets,sun397,UCF101,cheng2017resisc45}. We evaluate CLIP ResNet--50,
ResNet--101, ViT-B/32, and ViT-B/16. We sample 16 training examples
and at most four validation examples per class, and use batch size 64.
Following DVP, each downstream class has 20 fixed attribute descriptions,
making the source response dimension $20k_{\rm T}$. BDM is applied to the
complete temperature-scaled image-to-attribute-prompt similarity vector in
\Cref{eq:clip-attribute-logits}. For CLIP, we accumulate equivalent statistics in the normalized
visual-feature space and exploit an exact low-rank identity, avoiding
construction of the full logit covariance. YORO-FP uses a prompt learning rate of 0.1. Its
deterministic transform preserves aspect ratio before a $192\times192$ center
crop and the same padding. YORO-FP uses the same weak crop-and-flip family as the
full-data protocol before CLIP normalization.

\subsection{Dataset coverage}
\label[appendix]{app:datasets}

\begin{center}
\centering
\captionof{table}{Dataset coverage across the evaluated tasks and domains.
Full and FS indicate evaluation under the full-data and few-shot settings used in this work,
respectively.}
\label{tab:dataset-coverage}
\small
\setlength{\tabcolsep}{4.0pt}
\renewcommand{\arraystretch}{1.08}
\resizebox{0.7\linewidth}{!}{%
\begin{tabular}{l l c c}
\toprule
Dataset & Task/domain & Full & FS \\
\midrule
Flowers102 \citep{flowers102} & Fine-grained flowers & \checkmark & \checkmark \\
DTD \citep{DTD} & Texture attributes & \checkmark & \checkmark \\
UCF101 \citep{UCF101} & Human actions (video clips) & \checkmark & \checkmark \\
Food101 \citep{Food101} & Food recognition & \checkmark & \checkmark \\
GTSRB \citep{GTSRB} & Traffic signs & \checkmark & \checkmark \\
EuroSAT \citep{EuroSAT} & Satellite land use & \checkmark & \checkmark \\
Oxford-IIIT Pets \citep{Oxfordpets} & Fine-grained pets & \checkmark & \checkmark \\
Stanford Cars \citep{stanfordcars} & Fine-grained cars & \checkmark & \checkmark \\
SUN397 \citep{sun397} & Scene recognition & \checkmark & \checkmark \\
CIFAR-10 \citep{cifar10} & Natural images & \checkmark & \checkmark \\
CIFAR-100 \citep{cifar10} & Natural images & \checkmark & \checkmark \\
SVHN \citep{SVHN} & Street-view digits & \checkmark & \\
Caltech101 \citep{fei2004caltech101} & General objects & \checkmark & \checkmark \\
RESISC45 \citep{cheng2017resisc45} & Remote-sensing scenes & \checkmark & \checkmark \\
FGVC Aircraft \citep{maji2013aircraft} & Fine-grained aircraft & & \checkmark \\
\bottomrule
\end{tabular}}
\vspace{-0.1cm}
\end{center}

\paragraph{Full-data suite.}
Following BLM and BLM+ \citep{cai2024bayesian}, the principal comparison uses
the first 12 datasets in \Cref{tab:dataset-coverage}. They span
fine-grained objects, textures, actions, scenes, remote sensing, traffic signs,
and digits. The training-dynamics and mapping-update analyses add Caltech101
and RESISC45, giving the 14-task ablation suite. These two datasets are not
part of the published-baseline comparison.

\paragraph{Few-shot CLIP suite.}
Following DVP \citep{cai2025understanding}, the few-shot study uses the 11
datasets whose split column contains ``FS''. This suite deliberately combines fine-grained
recognition with textures, scenes, actions, and remote sensing. We keep DVP's
20 attribute descriptions per class fixed and treat the complete CLIP
image--text similarity vector as BDM's response. Each class contributes 16
training examples and at most four validation examples, allowing us to test
whether the same one-pass statistics remain useful when labels are scarce.

\paragraph{Model selection and evaluation.}
For each seed, we select
$\rho\in\{0.01,0.03,0.1,0.3,0.5,0.8\}$ on the validation split. The training
statistics and validation responses are computed once and shared by all six
candidates. Selection therefore adds only six analytic mapping solves and score
evaluations, with no additional training traversal or frozen-model evaluation.
Ties choose the smaller $\rho$. Official validation splits are used when available. CIFAR-10,
CIFAR-100, GTSRB, and SVHN use a stratified 10\% training holdout. YORO-FP
evaluates the unrefined YORO predictor together with refinement epochs 1--20,
then keeps the first checkpoint with the best validation accuracy. Test data
are evaluated once after selection.

BDM uses empirical class priors and solves the regularized linear system directly.
A small adaptive diagonal shift is added only when required for numerical
positive definiteness. Training time excludes validation and test
evaluation. YORO requires one labeled training-set traversal and no
backward pass, whereas YORO-FP uses 21 traversals in total. All main-text
results are macro-averaged across datasets and backbones.

\begin{algorithm}[H]
\footnotesize
\caption{Unified YORO and YORO-FP pseudocode. Within each minibatch, the
pre-update response is reused for both BDM statistics and prompt optimization,
so epoch-boundary mapping updates require no extra traversal.}
\label{alg:yoro}
\begin{algorithmic}[1]
\Require $\mathcal D_{\rm tr}$, $\mathcal D_{\rm val}$, frozen $f_{\rm pre}$,
$f_{\rm in}(\cdot\mid\theta_0)$, shrinkage $\rho$, epochs $E$, momentum $\beta$
\Ensure Validation-best $(\theta^\star,\omega^\star,b^\star)$
\State $\theta\gets\theta_0$; reset accumulator $\mathcal R$
\ForAll{$(X,Y)\in\mathcal D_{\rm tr}$} \Comment{one traversal}
  \State $Z\gets f_{\rm pre}(f_{\rm in}(X\mid\theta))$
  \State $\mathcal R\gets\Call{UpdateStats}{\mathcal R,Z,Y}$
\EndFor
\State $(\omega,b)\gets\Call{BDM}{\mathcal R,\rho}$; save the unrefined YORO candidate
\If{$E=0$}
  \State \Return the saved candidate \Comment{YORO}
\EndIf
\For{$t=0,\ldots,E-1$}
  \State reset $\mathcal R_t$
  \ForAll{$(X,Y)\in\mathcal D_{\rm tr}$}
    \State $Z\gets f_{\rm pre}(f_{\rm in}(X\mid\theta))$
    \State $\mathcal R_t\gets\Call{UpdateStats}{\mathcal R_t,
      \operatorname{stopgrad}(Z),Y}$
    \State $\theta\gets\Call{AdamStep}{\theta,
      \nabla_\theta\ell_{\rm CE}(\omega^\top Z+b,Y)}$
  \EndFor
  \State $(\widehat\omega,\widehat b)\gets\Call{BDM}{\mathcal R_t,\rho}$
  \State $(\omega,b)\gets\beta(\omega,b)
    +(1-\beta)(\widehat\omega,\widehat b)$
  \State save $(\theta,\omega,b)$ if validation accuracy strictly improves
\EndFor
\State \Return the saved checkpoint \Comment{YORO-FP}
\end{algorithmic}
\end{algorithm}

\subsection{Visual Prompt Configurations}
\label{sec:visual-prompt-config}

\paragraph{Padding and watermarking.}
Visual reprogramming modifies the input space using a trainable visual
pattern while keeping the pretrained backbone frozen. Following the
experimental settings of \citet{cai2025attribute,cai2025understanding},
we distinguish two common prompt geometries: padding-based and
watermarking-based visual prompts. A padding-based prompt first resizes the
downstream image to a smaller spatial resolution and places it at the center
of the pretrained model's input canvas. The remaining border is occupied by
trainable parameters. In contrast, a watermarking-based prompt resizes the downstream image to
the pretrained model's full input resolution and directly overlays the
trainable pattern within the image support. Thus, padding introduces
trainable parameters outside the resized image, whereas watermarking
directly perturbs the resized image itself.

Following the notation used in the visual reprogramming literature, we use
\emph{VP} to denote the visual-prompting baseline based on
\citet{bahng2022exploring}, which uses a watermarking-style prompt, and
\emph{AR} to denote adversarial reprogramming, which uses a padding-style
prompt. Although both approaches optimize input-space visual patterns, their
prompt widths have different meanings. For watermarking, the width specifies
the region of the original image on which the learned pattern is
superimposed. For padding, the width specifies the trainable border created
after resizing the downstream image.

\paragraph{Prompt size.}
Unless otherwise stated, our padding-based experiments use a 16-pixel-wide
visual prompt on each side, following DVP
\citep{cai2025understanding}. For a $224\times224$ input, the downstream
image is resized to $192\times192$ and centered in the input, leaving a
16-pixel trainable frame on each side. This resize-and-padding design also
reduces the number of trainable input parameters, since only the surrounding
frame is optimized while the resized image region remains fixed. We use this
configuration as the default padding setting throughout our experiments.

The reported VP and AR results follow the corresponding experimental
settings in \citet{cai2025attribute,cai2025understanding}, rather than our
default 16-pixel configuration. In particular, VP uses a watermarking prompt,
whose width specifies the region overlaid on the original image, whereas AR
uses a padding prompt, whose width specifies the border surrounding the
resized image. These values are therefore not directly comparable as prompt
widths. More generally, the resize-and-padding configuration can use fewer
trainable input parameters than a wider watermarking region because the
central resized image does not contain trainable prompt parameters.
Differences in prompt geometry and size across VP, AR, and our default
16-pixel padding setting follow the experimental protocols of the
corresponding methods.

\paragraph{Results under legacy padding.}
Table~\ref{tab:L-padding-detailed} further evaluates YORO under the legacy
padding protocol used by \citet{cai2024bayesian}. Unlike the visual prompt
configurations adopted in later work
\citep{cai2025attribute,cai2025understanding}, this protocol does not first
resize downstream images to a common spatial resolution. Instead, the
original downstream image is directly placed at the center of the
pretrained input canvas and the remaining area is filled by the visual
program. Consequently, the relative area occupied by the image depends on
its native resolution. For low-resolution datasets, only a small central
region may contain image content, producing a substantial mismatch with the
spatial scale seen by the pretrained model.

Later visual reprogramming methods have largely moved away from this legacy
configuration and instead operate on rescaled downstream images
\citep{cai2025attribute,cai2025understanding}. Rescaling provides a more
controlled input interface because downstream images occupy a consistent
fraction of the pretrained input across datasets. This distinction also
helps explain why YORO performs slightly better under our watermark setting:
the resized input provides a more informative and spatially consistent
source response, whereas legacy padding can substantially reduce the
effective image content for low-resolution datasets. Since YORO itself
requires no prompt training, its prediction is determined directly by the
source response induced by the given visual program and the proposed output
mapping.

Importantly, YORO remains strong even under this legacy padding protocol.
With ResNet--18, YORO achieves 54.7\% average accuracy, compared with
45.3\% for BLM and 46.7\% for BLM+, corresponding to gains of 9.4 and
8.0 percentage points, respectively. With ResNeXt--101--32x8d, YORO reaches
67.4\%, improving over BLM and BLM+ by 14.2 and 13.6 points. These gains are
obtained without iterative prompt optimization, even though the legacy
padding scheme does not resize downstream images to a common spatial scale
and can therefore provide a less consistent source response across datasets.
The improvement under the same input parameterization used by BLM indicates
that YORO's advantage primarily comes from its stronger use of the pretrained
model's output response rather than from a more favorable visual prompt
geometry. Together with the results under our default resized visual prompt
settings, this also shows that YORO is effective across different input
reprogramming configurations.

 \begin{table}[t]
\captionsetup{position=top}
\caption{\textbf{Comparison under the legacy padding setting}
(top-1 accuracy, \%). Following \citet{cai2024bayesian}, downstream images
are directly padded to the pretrained input size without resizing. Despite
this legacy input configuration, YORO substantially improves over BLM and
BLM+ across both pretrained backbones.}
\vspace{-0.2cm}
\label{tab:L-padding-detailed}
\centering
\small
\resizebox{\textwidth}{!}{%
\begin{tabular}{
c|cccccc|
>{\columncolor{blue!10}}c >{\columncolor{blue!20}}c|
ccccc|
>{\columncolor{blue!10}}c >{\columncolor{blue!20}}c
}
\toprule
& \multicolumn{8}{c|}{ResNet--18 (ImageNet--1K)}
& \multicolumn{7}{c}{ResNeXt--101--32x8d (Instagram)} \\
\midrule
Dataset & RLM & FLM & ILM & BLM & BLM+ & {\color{gray!70}Deep}
& YORO & YORO-FP & FLM & ILM & BLM & BLM+
& {\color{gray!70}Deep} & YORO & YORO-FP \\
\midrule
Flowers102 & 11.0 & 20.0 & 27.9 & 44.4 & 50.1 & {\color{gray!70}76.7} & 80.1 & \textbf{80.7} & 22.5 & 27.9 & 31.5 & 30.1 & {\color{gray!70}85.2} & 89.3 & \textbf{89.4} \\
DTD & 16.3 & 32.4 & 35.3 & 42.0 & 43.9 & {\color{gray!70}49.1} & \textbf{55.5} & 53.2 & 40.3 & 41.4 & 47.8 & 49.4 & {\color{gray!70}64.0} & \textbf{67.4} & 66.9 \\
UCF101 & 6.6 & 18.9 & 23.9 & 30.9 & 32.0 & {\color{gray!70}46.0} & \textbf{54.1} & 52.9 & 41.9 & 43.1 & 48.3 & 50.1 & {\color{gray!70}68.3} & 68.4 & \textbf{69.0} \\
Food101 & 3.8 & 12.8 & 14.8 & 23.2 & 25.1 & {\color{gray!70}34.1} & \textbf{41.7} & 35.5 & 20.5 & 23.0 & 29.6 & 31.4 & {\color{gray!70}58.7} & \textbf{60.9} & 58.2 \\
GTSRB & 46.1 & 45.5 & 52.0 & 54.8 & 54.3 & {\color{gray!70}63.1} & 47.2 & \textbf{57.7} & 56.2 & 59.9 & 62.9 & 63.0 & {\color{gray!70}74.4} & 58.2 & \textbf{67.9} \\
EuroSAT & 82.4 & 83.8 & 85.2 & 86.7 & 86.7 & {\color{gray!70}92.4} & 90.7 & \textbf{91.1} & 87.8 & 86.2 & 87.6 & 88.3 & {\color{gray!70}93.2} & 91.0 & \textbf{93.1} \\
OxfordPets & 9.3 & 62.9 & 65.4 & 69.8 & 70.6 & {\color{gray!70}73.0} & \textbf{77.0} & 76.6 & 76.8 & 78.9 & 82.4 & 83.0 & {\color{gray!70}91.8} & \textbf{92.7} & \textbf{92.7} \\
StanfordCars & 0.9 & 2.7 & 4.5 & 5.4 & 7.7 & {\color{gray!70}14.3} & 22.5 & \textbf{22.7} & 4.6 & 7.0 & 8.3 & 9.3 & {\color{gray!70}50.5} & \textbf{53.2} & 52.5 \\
SUN397 & 1.0 & 10.4 & 13.0 & 16.2 & 18.7 & {\color{gray!70}26.3} & \textbf{40.4} & 35.5 & 21.6 & 23.7 & 30.1 & 32.0 & {\color{gray!70}51.5} & \textbf{55.0} & 54.1 \\
CIFAR10 & 63.0 & 65.7 & 65.5 & 66.7 & 66.8 & {\color{gray!70}72.1} & 60.7 & \textbf{66.9} & 80.3 & 81.7 & \textbf{82.2} & \textbf{82.2} & {\color{gray!70}83.4} & 73.6 & 79.9 \\
CIFAR100 & 12.9 & 18.1 & 24.8 & 29.6 & 30.6 & {\color{gray!70}46.7} & 36.2 & \textbf{38.1} & 39.7 & 45.9 & 47.8 & 47.8 & {\color{gray!70}56.2} & 46.7 & \textbf{52.0} \\
SVHN & 73.5 & 73.1 & \textbf{75.2} & 74.5 & 74.2 & {\color{gray!70}82.1} & 50.8 & 72.9 & 79.0 & \textbf{81.4} & 79.8 & 79.3 & {\color{gray!70}85.7} & 52.4 & 81.2 \\
\midrule
Average & 27.2 & 37.2 & 40.6 & 45.3 & 46.7 & {\color{gray!70}56.3} & 54.7 & \textbf{57.0} & 47.6 & 50.0 & 53.2 & 53.8 & {\color{gray!70}71.9} & 67.4 & \textbf{71.4} \\
\bottomrule
\end{tabular}}
\end{table}

\section{Derivation and streaming equivalence}
\label[appendix]{app:proof}

\paragraph{Affine Bayes rule.}
Under \cref{eq:gaussian-model},
\[
\log p(z\mid c)+\log\pi_c
=-\tfrac12(z-\mu_c)^\top\Sigma^{-1}(z-\mu_c)+\log\pi_c+\mathrm{const}.
\]
Expanding the quadratic and dropping
$-\tfrac12z^\top\Sigma^{-1}z$, which is common to all classes, yields
the score in \cref{eq:lda-score} and the affine parameters
in \cref{eq:bdm-solve}.

\paragraph{Proof of \cref{prop:streaming-equivalence}.}
For class $c$, let $\widehat\mu_c=s_c/n_c$. Its within-class scatter is
\[
\sum_{i:y_i=c}(z_i-\widehat\mu_c)(z_i-\widehat\mu_c)^\top
=\sum_{i:y_i=c}z_iz_i^\top-\frac{s_cs_c^\top}{n_c}.
\]
Summing over classes gives
$Q-\sum_c s_cs_c^\top/n_c$, exactly the numerator
in \cref{eq:streaming-estimates}. Counts also recover the batch class means and
priors. Applying the same deterministic shrinkage and linear solve therefore
produces identical $(\omega_0,b_0)$. \hfill$\square$

\subsection{Optional exact low-rank CLIP implementation}
\label[appendix]{app:clip-low-rank}

BDM is defined entirely in the response space and does not require access to
intermediate model features. When CLIP visual features are available, however,
their low-dimensional structure can be exploited as an exact computational
shortcut. If only the response vectors are observable, BDM can instead be fit
directly from the response-space statistics in \cref{eq:streaming-statistics}.

Let $K=mk_{\rm T}$ and stack the normalized CLIP attribute embeddings as
$T\in\mathbb R^{K\times d}$. For a normalized visual feature
$v\in\mathbb R^d$, the response in \cref{eq:clip-attribute-logits} is
$z=\alpha Tv$, where $\alpha=\tau_{\rm C}$. Let
$M_v\in\mathbb R^{d\times k_{\rm T}}$ contain the target-class visual-feature
means and let $C_v\in\mathbb R^{d\times d}$ be their pooled within-class
covariance. The corresponding response-space statistics are
\begin{equation}
M_z=\alpha TM_v,
\qquad
\widehat\Sigma_z=\alpha^2TC_vT^\top,
\qquad
\tau=\frac{\alpha^2}{K}\operatorname{tr}(C_vT^\top T).
\label{eq:clip-low-rank-statistics}
\end{equation}

Writing $a=\rho\tau+\delta$ and $\beta=1-\rho$, the dense response-space BDM
system admits the exact solution
\begin{equation}
\omega=
\frac{\alpha}{a}T
\left(I_d+\frac{\beta\alpha^2}{a}C_vT^\top T\right)^{-1}M_v.
\label{eq:clip-low-rank-solve}
\end{equation}
This is a direct matrix-identity rewrite of
$(aI_K+\beta\alpha^2TC_vT^\top)\omega=\alpha TM_v$ and therefore produces
exactly the same affine mapping as dense response-space BDM, rather than a
low-rank approximation.

In our CLIP implementation, we exploit this identity by accumulating class
sums and the second moment in $d$ dimensions, solving the $d\times d$ system
in \cref{eq:clip-low-rank-solve}, and forming the final
$K\times k_{\rm T}$ mapping once. This use of visual features is purely an
implementation shortcut and does not change the BDM formulation.
Computation uses double-precision statistics, a symmetrized $C_v$, and the
adaptive diagonal floor $\delta$ from \cref{eq:shrinkage}.

\paragraph{Complexity.}
For generic classifier logits, accumulation costs
$\mathcal O(nk_{\rm S}^2)$ arithmetic and stores
$\mathcal O(k_{\rm S}^2+k_{\rm T}k_{\rm S})$ values; the final dense solve
costs $\mathcal O(k_{\rm S}^3+k_{\rm S}^2k_{\rm T})$. Both image-classifier
backbones in our experiments have $k_{\rm S}=1{,}000$, so the double-precision
second moment occupies approximately 7.6 MiB. For CLIP, the exact formulation
above stores $\mathcal O(d^2+k_{\rm T}d+Kd)$ values and solves a $d\times d$
system. The largest response in our suite is SUN397 with $K=20\times397=7{,}940$;
a dense $K\times K$ double-precision moment would contain 63.0 million entries
(approximately 481 MiB), whereas our implementation never materializes it.
The complete one-pass timings, including the analytic fit, are reported in
\cref{tab:compute-return}. All costs are independent of the number of prompt
optimization epochs because YORO has none.

\paragraph{Reproducibility.}
Our implementation keeps the backbone frozen and enforces one training-set
traversal for YORO. We will release the code, configurations, per-seed results,
and evaluation scripts.

\section{Detailed results and ablations}
\label[appendix]{app:detailed-results}

%All tables below report test accuracy. Validation data are used only for hyperparameter and checkpoint selection.

\begin{table}[t]
\caption{\textbf{Efficiency--accuracy comparison of YORO and YORO-FP.}
Accuracy is the three-seed macro-average over 14 full-data tasks for
ResNet--18 padding, or over 11 few-shot datasets and four CLIP backbones.
Relative time is normalized by the corresponding YORO run. All timings use
one NVIDIA A100 and exclude validation and test evaluation. ``No
refinement'' reports the fraction of YORO-FP runs in which validation
selects the original YORO checkpoint.}
\label{tab:compute-return}
\centering
\small
\setlength{\tabcolsep}{4pt}
\resizebox{0.8\columnwidth}{!}{%
\begin{tabular}{llcccc}
\toprule
Protocol & Method & Accuracy (\%) & $\Delta$ (\%) & No refinement
& Passes / time \\
\midrule
ResNet18 (P) & YORO & \ymean{71.0}{0.0} & -- & -- & $1$ / $1.0\times$ \\
ResNet18 (P) & YORO-FP & \ybest{71.6}{0.1} & $+0.6$ & 54.8\%
& $21$ / $20.8\times$ \\
\midrule
FS-CLIP & YORO & \ybest{77.2}{0.1} & -- & -- & $1$ / $1.0\times$ \\
FS-CLIP & YORO-FP & \ymean{77.0}{0.3} & $-0.2$ & 58.3\%
& $21$ / $17.5\times$ \\
\bottomrule
\end{tabular}}
\end{table}

\paragraph{Compute and accuracy.}
\Cref{tab:compute-return} shows diminishing returns from prompt
optimization. YORO-FP uses 21 labeled passes and about $18$--$21\times$ the
training time of YORO. This additional computation raises the full-data mean
by $0.6\%$, but lowers the few-shot CLIP mean by $0.2\%$. Validation selects
the original YORO checkpoint in 54.8\% and 58.3\% of runs, respectively.
YORO is therefore a strong accuracy--efficiency default, while YORO-FP
provides an optional refinement when validation favors further prompt
training.

\subsection{Additional covariance and response-selection ablations}

The additional covariance and response-selection comparisons are summarized in
\cref{tab:covariance-selection-ablation}. These supplementary ablations follow
the same frozen-response protocol as the main FS-CLIP experiments.

\begin{table}[h]
\caption{\textbf{Covariance and response-selection ablations.}
Results report mean $\pm$ sample standard deviation over three seeds (\%),
averaged across 11 datasets under the 16-shot FS-CLIP protocol.
Overall further averages the four CLIP backbones.}
\label{tab:covariance-selection-ablation}
\centering
\small
\setlength{\tabcolsep}{3.5pt}
\begin{tabular}{lcccccc}
\toprule
Backbone & BDM & Conditional 95\% & Adaptive & OAS & RBLW & Empirical LDA \\
\midrule
ResNet--50
& $\mathbf{74.4}\pm\mathbf{0.2}$
& $74.1\pm0.3$
& $74.0\pm0.3$
& $73.3\pm0.2$
& $73.3\pm0.2$
& $69.2\pm0.2$ \\

ResNet--101
& $\mathbf{76.8}\pm\mathbf{0.2}$
& $76.6\pm0.3$
& $76.6\pm0.3$
& $76.1\pm0.2$
& $76.1\pm0.2$
& $72.5\pm0.4$ \\

ViT-B/32
& $\mathbf{76.2}\pm\mathbf{0.3}$
& $76.1\pm0.4$
& $76.1\pm0.4$
& $75.5\pm0.0$
& $75.5\pm0.0$
& $71.6\pm0.1$ \\

ViT-B/16
& $\mathbf{81.2}\pm\mathbf{0.1}$
& $81.1\pm0.1$
& $81.0\pm0.2$
& $80.7\pm0.1$
& $80.7\pm0.1$
& $77.5\pm0.5$ \\

\midrule
Overall
& $\mathbf{77.2}\pm\mathbf{0.1}$
& $77.0\pm0.0$
& $76.9\pm0.1$
& $76.4\pm0.1$
& $76.4\pm0.1$
& $72.7\pm0.2$ \\
\bottomrule
\end{tabular}
\end{table}

\paragraph{Covariance regularization is more important than response pruning.}
On FS-CLIP, empirical LDA, which uses the unregularized shared sample
covariance, reaches $72.7\%$, while default BDM reaches $77.2\%$. OAS and
RBLW \citep{chen2010shrinkage,ledoit2004well} also improve substantially over unregularized
LDA, reaching $76.4\%$ without validation-based selection of the shrinkage
coefficient.

Conditional 95\% uses training labels to greedily select response coordinates
according to their additional class-discriminative information conditioned on
the coordinates already selected, and retains a fixed 95\% of the response
dimensions. Adaptive instead jointly selects the retention ratio, including
the option to keep all coordinates, and the covariance-shrinkage coefficient
using validation data for each dataset, backbone, and seed. It retains
97.3\% of the coordinates on average, indicating that validation generally
favors keeping most of the response.

Neither strategy improves on full-response BDM. Conditional 95\% reaches
$77.0\%$, and Adaptive reaches $76.9\%$, compared with $77.2\%$ for BDM.
We therefore retain the complete response by default and use shrinkage to
stabilize covariance estimation rather than explicitly pruning response
coordinates.

\subsection{Contextual comparison with efficient CLIP adaptation}

\begin{table}[h]
\caption{\textbf{Computationally efficient CLIP adaptation on ten overlapping
datasets with a ResNet--50 backbone} (accuracy \%). We recompute each average
over OxfordPets, Flowers102, FGVC Aircraft, DTD, EuroSAT, StanfordCars,
Food101, SUN397, Caltech101, and UCF101. Tip-Adapter is reported by
\citet{wang2024hard}, while CARPRT, MPE, and WPE are reported by
\citet{dong2026carprt}. YORO is our three-seed result. All methods avoid
iterative backbone updates, so the table provides a contextual efficiency-oriented
comparison despite differences in data access.}
\label{tab:external-fs-clip}
\centering
\small
\setlength{\tabcolsep}{4.2pt}
\resizebox{\columnwidth}{!}{%
\begin{tabular}{lccccccccccc}
\toprule
Method & Pets & Flowers & Aircraft & DTD & EuroSAT & Cars & Food & SUN & Caltech & UCF & Avg. \\
\midrule
CARPRT & 85.7 & 65.6 & 16.9 & 41.3 & 36.8 & 56.4 & 76.9 & 61.3 & 88.5 & 63.7 & 59.3 \\
MPE & 76.0 & 57.0 & 16.1 & 41.7 & 30.3 & 55.7 & 75.5 & 59.3 & 86.4 & 60.1 & 55.8 \\
WPE & 78.4 & 58.8 & 16.1 & 40.9 & 30.7 & 56.0 & 76.2 & 59.7 & 86.7 & 61.5 & 56.5 \\
Tip-Adapter & \textbf{88.1} & 89.9 & 29.8 & 60.9 & 70.5 & 66.8 & \textbf{77.8} & 66.9 & 90.2 & 70.6 & 71.2 \\
\rowcolor{blue!10}
YORO & 87.5 & \textbf{92.6} & \textbf{36.2} & \textbf{64.4} & \textbf{82.9} & \textbf{68.6} & 72.8 & \textbf{67.6} & \textbf{91.3} & \textbf{75.1} & \textbf{73.9} \\
\bottomrule
\end{tabular}}
\end{table}

The broader comparison in \Cref{tab:external-fs-clip} positions YORO among
computationally efficient CLIP adaptation methods. On the ten-dataset overlap,
YORO exceeds Tip-Adapter by 2.8\% while using the complete
attribute-prompt response as its only source representation. This comparison
is intended to characterize efficiency-oriented alternatives rather than merge
incompatible training protocols. YORO's advantage comes from its discriminant
output mapping, not from an expensive feature adapter or prolonged backbone
training.

\begin{table}[H]
\caption{\textbf{SVHN isolates the benefit of input adaptation.} With BDM
frozen, optimizing only the visual prompt improves every tested
backbone--input setting. Accuracy is mean $\pm$ sample standard deviation over
three seeds (\%).}
\label{tab:svhn-prompt-only}
\centering
\small
\setlength{\tabcolsep}{5pt}
\begin{tabular}{llcccc}
\toprule
Backbone & Input & YORO & Frozen BDM & $\Delta$ (\%) & YORO-FP \\
\midrule
ResNet--18 & Padding
& \ymean{72.3}{0.1} & \ymean{76.0}{0.2} & $+3.7$ & \ybest{77.0}{0.3} \\
ResNeXt--101--32x8d & Padding
& \ymean{56.5}{0.0} & \ymean{63.3}{0.3} & $+6.8$ & \ybest{68.6}{0.6} \\
ResNet--18 & Watermark
& \ymean{60.6}{0.1} & \ymean{83.4}{0.2} & $+22.8$ & \ybest{84.7}{0.2} \\
\bottomrule
\end{tabular}
\end{table}

\paragraph{Why SVHN is different.}
With BDM frozen, prompt training raises SVHN accuracy by 3.7\%, 6.8\%, and
22.8\% across the three settings in \cref{tab:svhn-prompt-only}. The visual
prompt therefore supplies most of the improvement even when the initial label
mapping is held fixed. The particularly large watermarking gain shows that
the benefit cannot be explained by repeatedly refitting the output mapping:
input adaptation is reshaping the frozen response geometry itself. YORO-FP
adds another 1.0\%, 5.3\%, and 1.3\% by tracking that moving geometry without
replacing the mapping abruptly. Together with the aggregate result in
\cref{tab:compute-return}, SVHN identifies the exception that motivates
YORO-FP: prompt training is valuable when the source responses and target
classes are initially poorly aligned, rather than as a universal default.

\begin{table}[H]
\captionsetup{position=top}
\caption{\textbf{Variants of YORO-FP mapping strategies.}
$\Delta$ is measured relative to YORO, with 95\% bootstrap confidence
intervals over evaluation settings. W/T/L counts datasets for full data and
dataset--backbone pairs in the few-shot CLIP setting. Refinement depth is the number of additional prompt-training epochs selected on the validation set. A depth of 0 corresponds to the original YORO checkpoint. A median depth of 0 means that validation selects YORO without further prompt training in at least half of the runs. The last column
reports the fraction of runs for which validation retains the original YORO
predictor. Frozen, Refresh, and Momentum use $\beta=1$, $0$, and $0.9$,
respectively.}
\label{tab:yoro-mapping-diagnostics}
\centering
\small
\setlength{\tabcolsep}{3pt}
\begin{tabular}{llcccccc}
\toprule
Protocol & Strategy & Accuracy (\%) & $\Delta$ (\%) & 95\% CI
& W/T/L & Refinement depth & No refinement \\
\midrule
Full data & Frozen   & $71.1\pm0.0$ & $+0.2$ & $[-0.2,+0.8]$
& 2/7/5   & $4.8\pm7.7$ & 69.0\% \\
          & Refresh  & $71.1\pm0.0$ & $+0.1$ & $[-0.2,+0.7]$
& 2/10/2  & $2.2\pm5.5$ & 81.0\% \\
%\rowcolor{blue!20}
          & Momentum & $\mathbf{71.6\pm0.1}$ & $\mathbf{+0.6}$
& $\mathbf{[0.0,+1.4]}$ & 5/6/3 & $6.5\pm7.8$ & 54.8\% \\
\midrule
FS-CLIP   & Frozen   & $76.8\pm0.1$ & $-0.3$ & $[-0.5,-0.2]$
& 2/13/29 & $3.2\pm5.6$ & 66.7\% \\
          & Refresh  & $76.9\pm0.1$ & $-0.3$ & $[-0.5,0.0]$
& 6/14/24 & $3.4\pm5.7$ & 60.6\% \\
%\rowcolor{blue!20}
          & Momentum & $\mathbf{77.0\pm0.3}$ & $\mathbf{-0.2}$
& $\mathbf{[-0.4,0.0]}$ & 8/8/28 & $4.2\pm6.3$ & 58.3\% \\
\bottomrule
\end{tabular}
\end{table}

\paragraph{Selection behavior.}
\Cref{tab:yoro-mapping-diagnostics} separates the value of prompt refinement
from the way BDM follows the changing responses. In full data, Frozen and
Refresh improve the mean by only $0.2\%$ and $0.1\%$, whereas Momentum reaches
$+0.6\%$ and produces the largest number of wins. This supports gradual
tracking: an exponential moving average adapts the mapping when the prompt
changes the response geometry, while avoiding the instability of replacing it
after every epoch. The gain remains selective, however, because Momentum still
retains the original YORO predictor in 54.8\% of runs.

The few-shot regime is different. All three update rules reduce the aggregate
accuracy and losses outnumber wins, with no refinement selected in 58.3--66.7\%
of runs. Limited supervision leaves little evidence for reliably updating both
the prompt and its label mapping. Thus, the two protocols yield a consistent
decision rule: begin with the strong one-pass YORO predictor, and spend the
few-pass budget only when held-out validation demonstrates that response-space
adaptation transfers.

\begin{table}[p]
\caption{\textbf{Per-dataset full-data mapping-update ablation} (accuracy \%).
Each entry is the three-seed mean $\pm$ sample standard deviation. P/W denote
padding/watermarking. Bold marks the highest mean within each dataset and
setting. These results underlie the averages in
\cref{tab:yoro-mapping-full}.}
\label{tab:yoro-mapping-full-detail}
\centering
\footnotesize
\setlength{\tabcolsep}{7pt}
\renewcommand{\arraystretch}{0.84}
\begin{tabular}{@{}lcccc@{}}
\toprule
Dataset & YORO & Frozen & Refresh & Momentum \\
\midrule
\multicolumn{5}{l}{\textit{(a) ResNet--18 / P}} \\
Caltech101   & \abacc{92.0}{0.0} & \abacc{92.0}{0.0} & \abacc{92.0}{0.0} & \abest{92.1}{0.1} \\
CIFAR-10     & \abest{86.2}{0.1} & \abacc{86.0}{0.4} & \abest{86.2}{0.1} & \abest{86.2}{0.1} \\
CIFAR-100    & \abest{64.8}{0.1} & \abest{64.8}{0.1} & \abest{64.8}{0.1} & \abest{64.8}{0.1} \\
DTD          & \abacc{59.0}{0.0} & \abacc{59.0}{0.0} & \abacc{59.1}{0.2} & \abest{59.6}{0.5} \\
EuroSAT      & \abest{92.8}{0.0} & \abacc{92.5}{0.5} & \abacc{92.5}{0.5} & \abacc{92.5}{0.2} \\
Flowers102   & \abest{88.0}{0.0} & \abacc{87.8}{0.3} & \abest{88.0}{0.0} & \abacc{87.9}{0.3} \\
Food101      & \abest{49.8}{0.0} & \abest{49.8}{0.0} & \abest{49.8}{0.0} & \abest{49.8}{0.0} \\
GTSRB        & \abacc{79.0}{0.1} & \abacc{79.4}{0.3} & \abacc{79.0}{0.1} & \abest{80.2}{0.4} \\
OxfordPets   & \abest{86.9}{0.0} & \abest{86.9}{0.0} & \abest{86.9}{0.0} & \abest{86.9}{0.0} \\
RESISC45     & \abest{78.8}{0.0} & \abest{78.8}{0.0} & \abest{78.8}{0.0} & \abacc{78.7}{0.2} \\
StanfordCars & \abacc{34.9}{0.0} & \abacc{33.9}{0.2} & \abacc{33.9}{0.2} & \abest{37.1}{0.1} \\
SUN397       & \abest{47.8}{0.0} & \abest{47.8}{0.0} & \abest{47.8}{0.0} & \abest{47.8}{0.0} \\
SVHN         & \abacc{72.3}{0.1} & \abacc{76.0}{0.2} & \abacc{75.3}{0.4} & \abest{77.0}{0.3} \\
UCF101       & \abest{61.3}{0.0} & \abest{61.3}{0.0} & \abest{61.3}{0.0} & \abest{61.3}{0.0} \\
\addlinespace[1pt]
Average      & \abacc{71.0}{0.0} & \abacc{71.1}{0.0} & \abacc{71.1}{0.0} & \abest{71.6}{0.1} \\
\midrule
\multicolumn{5}{l}{\textit{(b) ResNet--18 / W}} \\
Caltech101   & \abest{92.4}{0.0} & \abest{92.4}{0.0} & \abest{92.4}{0.0} & \abest{92.4}{0.0} \\
CIFAR-10     & \abest{86.9}{0.1} & \abest{86.9}{0.1} & \abest{86.9}{0.1} & \abest{86.9}{0.1} \\
CIFAR-100    & \abest{64.8}{0.1} & \abest{64.8}{0.1} & \abest{64.8}{0.1} & \abest{64.8}{0.1} \\
DTD          & \abest{62.5}{0.0} & \abest{62.5}{0.0} & \abacc{61.3}{1.2} & \abest{62.5}{0.0} \\
EuroSAT      & \abest{93.2}{0.0} & \abest{93.2}{0.0} & \abest{93.2}{0.0} & \abest{93.2}{0.0} \\
Flowers102   & \abest{88.8}{0.0} & \abest{88.8}{0.0} & \abest{88.8}{0.0} & \abest{88.8}{0.0} \\
Food101      & \abest{52.8}{0.0} & \abest{52.8}{0.0} & \abest{52.8}{0.0} & \abest{52.8}{0.0} \\
GTSRB        & \abacc{75.9}{0.1} & \abacc{79.5}{0.6} & \abest{85.2}{0.9} & \abacc{84.5}{0.6} \\
OxfordPets   & \abest{88.5}{0.0} & \abacc{88.1}{0.6} & \abest{88.5}{0.0} & \abest{88.5}{0.0} \\
RESISC45     & \abest{79.5}{0.0} & \abest{79.5}{0.0} & \abest{79.5}{0.0} & \abest{79.5}{0.0} \\
StanfordCars & \abacc{37.8}{0.0} & \abacc{37.8}{0.0} & \abacc{37.8}{0.0} & \abest{38.6}{0.5} \\
SUN397       & \abest{52.0}{0.0} & \abest{52.0}{0.0} & \abest{52.0}{0.0} & \abest{52.0}{0.0} \\
SVHN         & \abacc{60.6}{0.1} & \abacc{83.4}{0.2} & \abacc{82.3}{0.4} & \abest{84.7}{0.2} \\
UCF101       & \abest{65.1}{0.0} & \abest{65.1}{0.0} & \abest{65.1}{0.0} & \abest{65.1}{0.0} \\
\addlinespace[1pt]
Average      & \abacc{71.5}{0.0} & \abacc{73.3}{0.1} & \abacc{73.6}{0.1} & \abest{73.9}{0.0} \\
\midrule
\multicolumn{5}{l}{\textit{(c) ResNeXt--101--32x8d / P}} \\
Caltech101   & \abest{96.5}{0.0} & \abacc{96.2}{0.1} & \abacc{96.4}{0.2} & \abacc{96.4}{0.0} \\
CIFAR-10     & \abest{95.3}{0.0} & \abest{95.3}{0.0} & \abest{95.3}{0.0} & \abest{95.3}{0.0} \\
CIFAR-100    & \abest{76.3}{0.1} & \abest{76.3}{0.1} & \abest{76.3}{0.1} & \abest{76.3}{0.1} \\
DTD          & \abest{70.3}{0.0} & \abest{70.3}{0.0} & \abest{70.3}{0.0} & \abest{70.3}{0.0} \\
EuroSAT      & \abacc{93.0}{0.0} & \abacc{92.8}{0.4} & \abacc{93.0}{0.2} & \abest{93.4}{0.3} \\
Flowers102   & \abacc{94.0}{0.0} & \abacc{94.0}{0.0} & \abacc{94.0}{0.0} & \abest{94.3}{0.1} \\
Food101      & \abest{70.3}{0.0} & \abest{70.3}{0.0} & \abest{70.3}{0.0} & \abest{70.3}{0.0} \\
GTSRB        & \abacc{79.3}{0.1} & \abacc{79.3}{0.1} & \abacc{79.3}{0.1} & \abest{79.6}{0.5} \\
OxfordPets   & \abacc{94.8}{0.0} & \abacc{94.9}{0.1} & \abest{95.2}{0.1} & \abacc{95.1}{0.1} \\
RESISC45     & \abacc{82.5}{0.0} & \abacc{82.3}{0.1} & \abacc{82.5}{0.0} & \abest{83.2}{0.3} \\
StanfordCars & \abacc{71.1}{0.0} & \abacc{70.9}{0.2} & \abacc{70.0}{0.2} & \abest{72.2}{0.3} \\
SUN397       & \abest{61.9}{0.0} & \abacc{61.8}{0.2} & \abest{61.9}{0.0} & \abest{61.9}{0.0} \\
SVHN         & \abacc{56.5}{0.0} & \abacc{63.3}{0.3} & \abacc{68.1}{0.7} & \abest{68.6}{0.6} \\
UCF101       & \abacc{74.9}{0.0} & \abacc{74.8}{0.1} & \abacc{75.1}{0.2} & \abest{75.4}{0.3} \\
\addlinespace[1pt]
Average      & \abacc{79.8}{0.0} & \abacc{80.2}{0.0} & \abacc{80.6}{0.1} & \abest{80.9}{0.1} \\
\bottomrule
\end{tabular}
\end{table}

\newpage

\section{Notation}
\label[appendix]{app:notation}

\Cref{tab:notation,tab:notation-cont} list the symbols used in the method and
result tables.
Superscripts ${\rm S}$ and ${\rm T}$ always mean source and target.

\begin{table}[H]
\caption{Symbols used in YORO and YORO-FP.}
\label{tab:notation}
\centering
\small
\def\arraystretch{1.35}
\centerline{\bfseries Spaces and Indexing}
\bgroup
\begin{tabular}{@{}p{1.25in}p{4.00in}@{}}
$\mathcal X^{\rm S},\mathcal X^{\rm T}$ & Source and target input spaces.\\
$\mathcal Y^{\rm S},\mathcal Y^{\rm T}$ & Source and target label spaces.\\
$k_{\rm S}$ & Number of source-response coordinates.\\
$k_{\rm T}$ & Number of target classes.\\
$[q]$ & Index set $\{1,\ldots,q\}$.\\
$i$ & Example index.\\
$c$ & Target-class index.\\
$j$ & Attribute-prompt index.\\
$\mathbb R^d$ & Real vector space of dimension $d$.\\
$\langle\cdot,\cdot\rangle$ & Inner product.\\
$(x_i^{\rm T},y_i^{\rm T})$ & Target example and label.\\
$n$ & Sample count.\\
$\mathcal D_{\rm tr},\mathcal D_{\rm val}$ & Training and validation sets.\\
\end{tabular}
\egroup
\vspace{0.16cm}

\centerline{\bfseries BDM and Output Mapping}
\bgroup
\begin{tabular}{@{}p{1.25in}p{4.00in}@{}}
$\mu_c$ & Class mean for target class $c$.\\
$\pi_c$ & Class prior for target class $c$.\\
$\Sigma$ & Shared response covariance.\\
$\mathcal N(\mu,\Sigma)$ & Gaussian distribution with mean $\mu$ and covariance $\Sigma$.\\
$p(\cdot)$ & Probability model.\\
$\widehat\mu_c$ & One-pass estimate of the class mean.\\
$\widehat\pi_c$ & One-pass estimate of the class prior.\\
$\widehat\Sigma$ & Empirical covariance estimate.\\
$\rho$ & Covariance shrinkage coefficient.\\
$\delta$ & Numerical diagonal shift.\\
$\widehat\Sigma_\rho$ & Regularized covariance used by BDM.\\
$I$ & Identity matrix.\\
$\operatorname{tr}(\cdot)$ & Matrix trace.\\
$\mathbf 1[\cdot]$ & Indicator function.\\
$M$ & Class-mean matrix $[\widehat\mu_1,\ldots,\widehat\mu_{k_{\rm T}}]$.\\
$\omega$ & Affine label-mapping weights.\\
$b$ & Affine label-mapping bias.\\
$f_{\rm out}^{\omega,b}$ & Affine output mapping.\\
$g_c$ & BDM score for class $c$.\\
$\widehat\omega,\widehat b$ & BDM candidate from the current pass. Subscript 0 marks the first pass.\\
$n_c$ & Number of responses assigned to class $c$.\\
$s_c$ & Sum of responses assigned to class $c$.\\
$S$ & Class-sum matrix.\\
$Q$ & Global uncentered second moment.\\
$\mathcal R,\mathcal R_t$ & Running BDM statistics $(\mathbf n,S,Q)$ for the first pass and epoch $t$.\\
\end{tabular}
\egroup
\end{table}

\newpage
\begin{table}[H]
\captionsetup{position=top}
\caption{Symbols used in YORO and YORO-FP (continued).}
\label{tab:notation-cont}
\centering
\small
\def\arraystretch{1.35}

\vspace{0.16cm}

\centerline{\bfseries Reprogramming and Optimization}
\bgroup
\begin{tabular}{@{}p{1.25in}p{4.00in}@{}}
$f_{\rm pre}$ & Frozen classifier or CLIP encoders.\\
$f_{\rm in}(\cdot\mid\theta)$ & Input reprogramming function with parameters $\theta$.\\
$\theta_0$ & Prompt initialization.\\
$\mathcal L_{\rm FP}$ & YORO-FP objective.\\
$\ell_{\rm CE}$ & Cross-entropy loss.\\
$\nabla_\theta$ & Gradient with respect to prompt parameters.\\
$X$ & Minibatch images.\\
$Y$ & Minibatch labels.\\
$Z$ & Minibatch response vectors.\\
$z_i(\theta)$ & Response vector for example $i$.\\
$\operatorname{stopgrad}(\cdot)$ & Blocks gradients through its argument.\\
$\textsc{UpdateStats}$ & Updates the streaming class statistics.\\
$\textsc{BDM}$ & Analytic discriminant mapping fit.\\
$\textsc{AdamStep}$ & One Adam optimizer update of the prompt.\\
$E$ & YORO-FP epoch budget.\\
$t$ & Epoch index.\\
$\beta$ & Momentum weight.\\
$(\cdot)^\star$ & Validation-selected checkpoint.\\
\end{tabular}
\egroup
\vspace{0.16cm}

\centerline{\bfseries CLIP and Evaluation}
\bgroup
\begin{tabular}{@{}p{1.25in}p{4.00in}@{}}
$\mathcal A$ & CLIP attribute-prompt bank.\\
$a_{c,j}$ & Prompt $j$ for class $c$.\\
$m$ & Number of prompts per class.\\
$\phi_{\rm I}$ & Frozen CLIP image encoder.\\
$\phi_{\rm T}$ & Frozen CLIP text encoder.\\
$\bar\phi$ & Normalized CLIP output.\\
$\tau_{\rm C}$ & Frozen CLIP logit scale.\\
$\Delta$ & Accuracy change in percent.\\
$\mathrm{CI}$ & Confidence interval.\\
$\mathrm{W/T/L}$ & Win, tie, and loss counts.\\
\end{tabular}
\egroup
\end{table}

\end{document}